\documentclass[draftclsnofoot, 11pt,onecolumn]{IEEEtran}
\usepackage{amsmath,epsfig}
\usepackage{verbatim} 
\usepackage{bm}
\usepackage{subfigure}
\usepackage{epsfig}
\usepackage{stmaryrd}
\usepackage[ruled,vlined]{algorithm2e}
\usepackage{amssymb}
\usepackage{textcomp}
\usepackage{wasysym}

\def\l|{\left|\left|}
\def\r|{\right|\right|}
\def\(|{\left(}
\def\)|{\right)}

\def\diag{\mbox{diag}}
\def\mod{\mbox{mod}}
\def\vec{\mbox{vec}}

\def\trace{\mbox{tr}}
\def\mod{\mbox{mod}}
\renewcommand{\b}{\bm}

\title{ Order-preserving factor analysis (OPFA)}
\author{ Arnau Tibau Puig and  Alfred O. Hero III}

\begin{document}

\vspace{150pt}
\begin{center}
\huge \textbf{Order-preserving factor analysis (OPFA)} \\
\vspace{20pt}
\large  Arnau Tibau Puig and  Alfred O. Hero III \\
\vspace{20pt}
\large EECS Department, University of Michigan, Ann Arbor, MI 48109-2122, USA 
\end{center}

\vspace{50pt}
\begin{center}
\textbf{Communications and Signal Processing Laboratory} \\
{Technical Report}: cspl-396 \\
{Date}: May. 5 2011 (v2) \\
A shorter version of this TR was accepted for publication to the IEEE Transactions on Signal Processing in April 2011.
\end{center}

\newpage  

\maketitle

\section{Introduction}
\label{sec:intro}

With the advent of high-throughput data collection techniques,
low-dimensional matrix factorizations have become an essential tool
for pre-processing, interpreting or compressing high-dimensional
data. They are widely used in a variety of signal processing domains
including electrocardiogram \cite{johnstone2007sparse}, image
\cite{jenatton2009structured}, or sound \cite{blumensath2006sparse}
processing. These methods can take advantage of a large range of a
priori knowledge on the form of the factors, enforcing it
 through constraints on sparsity or patterns in the
factors. However, these methods do not work well when there are
unknown misalignments between subjects in the population, e.g.,
unknown subject-specific time shifts. In such cases, one cannot apply
standard patterning constraints without first aligning the data; a
difficult task. An alternative approach, explored in this paper, is to
impose a factorization constraint that is invariant to factor
misalignments but preserves the relative ordering of the factors over
the population.  This order-preserving factor analysis is accomplished
using a penalized least squares formulation using shift-invariant yet
order-preserving model selection (group lasso) penalties on the
factorization. As a byproduct the factorization produces estimates of
the factor ordering and the order-preserving time shifts.

In traditional matrix factorization, the data is modeled
as a linear combination of a number of factors. Thus, given an $n
\times p$ data matrix $\b{X}$, the Linear Factor model is defined
as:
\begin{eqnarray}
\label{simplefamodel}
\b{X} = \b{M} \b{A} + \b{\epsilon},
\end{eqnarray}
where $\b{M}$ is a $n \times f$ matrix of factor loadings or
dictionary elements, $\b{A}$ is a $f \times p$ matrix of scores (also
called coordinates) and $\epsilon$ is a small residual. For example,
in a gene expression time course analysis, $n$ is the number of time points and $p$ is the number of genes in the study, 
 the columns of $\b{M}$ contain the features summarizing the genes' temporal trajectories
and the columns of $\b{A}$ represent the coordinates of each gene on the space
spanned by $\b{M}$.  Given this
model, the problem is to find a parsimonious factorization that fits
the data well according to selected criteria, e.g. minimizing the
reconstruction error or maximizing the explained variance. There are
two main approaches to such a parsimonious factorization. One, called
Factor Analysis, assumes that the number of factors is small and
yields a low-rank matrix factorization \cite{pearson1901liii},
\cite{carroll1970analysis}. The other, called Dictionary Learning
\cite{aharon2006k}, \cite{kreutz2003dictionary} or Sparse Coding
\cite{olshausen1997sparse}, assumes that the loading matrix $\b{M}$
comes from an overcomplete dictionary of functions and results in a
sparse score matrix $\b{A}$. There are also hybrid approaches such as
Sparse Factor Analysis \cite{johnstone2007sparse},
\cite{witten2009penalized}, \cite{jenatton2009structured} that try to
enforce low rank and sparsity simultaneously.

In many situations, we observe not one but several matrices $\b{X}_s$, $s=1,\cdots,S$ 
and there are physical grounds for believing that the $\b{X}_s$'s
share an underlying model. This happens, for instance, when the
observations consist of different time-blocks of sound from the same
music piece \cite{blumensath2006sparse}, \cite{mailhe2008shift}, when
they consist of time samples of gene expression microarray data from
different individuals inoculated with the same virus \cite{zaas2009gene},
or when they arise from the reception of digital data with code,
spatial and temporal diversity \cite{sidiropoulos2000blind}. In these
situations, the fixed factor model (\ref{simplefamodel}) is overly
simplistic.

\begin{figure}[h!]
\centerline{\includegraphics[trim = 15mm 1mm 15mm 5mm, clip,width=250pt]{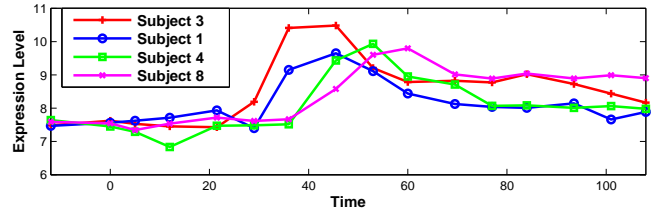}}
\caption{ Example of temporal misalignment across subjects of upregulated gene
  \textit{CCRL2}. Subject 6 and subject 10 show the earliest and the
  latest up-regulation responses, respectively.}
\label{Fig:misalignment}
\end{figure}

An example, which is the main motivation for this work is shown in
Figure \ref{Fig:misalignment}, which shows the effect of temporal
misalignment across subjects in a viral challenge study reported in 
\cite{zaas2009gene}. Figure \ref{Fig:misalignment} shows the expression trajectory
 for a particular gene that undergoes an increase (up-regulation) after viral inoculation at time 0, where the  
 moment when up-regulation occurs differs
over the population. Training the model (\ref{simplefamodel}) on this
data will produce poor fit due to misalignment of gene expression
onset times.

A more sensible approach for the data in Figure \ref{Fig:misalignment}
would be to separately fit each subject with a translated version of a
common up-regulation factor. This motivates the following extension of
model (\ref{simplefamodel}), where the factor matrices $\b{M}_s$,
$\b{A}_s$ are allowed to vary across observations. Given a number $S$ of $n
\times p$ data matrices $\b{X}_s$, we let:
\begin{eqnarray}
\label{originalstructure}
\b{X}_s = \b{M}_s \b{A}_s + \b{\epsilon}_s & s=1,\cdots,S.
\end{eqnarray}
Following the gene expression example, here $n$ is the number of time points, $p$ is the number of genes in the study, 
and $S$ is the number of subjects participating in the study. 
Hence, the $n \times f$ matrices $\b{M}_s$ contain the translated temporal features corresponding to the $s$-th subject and 
the $f \times p$ matrices $\b{A}_s$ accommodate
the possibility of subjects having different mixing weights.
 For different constraints on $\b{M}_s$, $\b{A}_s$, this model
specializes to several well-known paradigms such as Principal
Components Analysis (PCA) \cite{pearson1901liii}, sparse PCA
\cite{johnstone2007sparse}, k-SVD \cite{aharon2006k}, structured PCA
\cite{jenatton2009structured}, Non-Negative Matrix Factorization
(NNMF) \cite{lee1999learning}, Maximum-Margin Matrix Factorization (MMMF) \cite{srebro2005maximum},
Sparse Shift-invariant models \cite{blumensath2006sparse}, Parallel Factor Analysis (PARAFAC)
\cite{carroll1970analysis}, \cite{kolda2009tensor} or Higher-Order SVD
(HOSVD) \cite{Larsson2010HOSVD}. Table \ref{Table:review} summarizes the characteristics of these decomposition models when seen as different instances of the general model (\ref{originalstructure}).

\begin{table}[htb]
\caption{\label{Table:review} Special cases of the general model (\ref{originalstructure}). } 
\begin{center}
\footnotesize 
\begin{tabular}{|c|c|c|c|c|c|c|}
  \hline
   \textbf{Decomposition} & \textbf{Structure of $\b{M}_s$} & \textbf{Structure of $\b{A}_s$} &  \textbf{Uniqueness} &  \textbf{Reference}\\
  \hline
   PCA & Orthogonal $\b{M}_s=\b{F}$  & Orthogonal $\b{A}_s$& \tiny  Yes & SVD  \\
  \hline
   Sparse-PCA & Sparse $\b{M}_s=\b{F}$ & Sparse $\b{A}_s$ & \tiny No & Sparse PCA \cite{johnstone2007sparse}, \cite{d2007direct}, \\
    &  &  &   &  k-SVD \cite{aharon2006k}, PMD \cite{witten2009penalized} \\
  \hline
   Structured-PCA & $\b{M}_s=\b{F}$ & Structured Sparse $\b{A}_s$ & \tiny No & \cite{jenatton2009structured}\\
  \hline
   NNMF & Non-negative $\b{M}_s=\b{F}$ & Non-negative $\b{A}_s$ & \tiny No & \cite{lee1999learning}\\
\hline
Sparse   & $\b{M}_s=\left[\b{M}\left(\b{F},d_1\right) \cdots \b{M}\left(\b{F},d_D\right)\right]$ & Sparse $\b{A}_s$  & \tiny No & \cite{lewicki1999coding},\\
Shift-invariant & where $\left\{d_j\right\}_{j=1}^D$ are all possible & & & \cite{blumensath2006sparse}, \cite{mailhe2008shift} \\
models & translations of the $n$-dimensional vectors in $\b{F}$. & & & \\
  \hline
   PARAFAC/CP & $\b{M}_s=\b{F}$ & $\b{A}_s=\diag\left(\b{C}_{\cdot,s}\right) \b{B}'$ & \tiny Yes & \cite{kolda2009tensor}\\
  \hline
   HOSVD & Orthogonal $\b{M}_s=\b{F}$ & $\b{A}_s=\left(\mathcal{G} \times_{3} \b{C}_{\cdot,s} \right) \b{B}'$ & \tiny Yes & \cite{Larsson2010HOSVD} \\
   &   & where slices of $\mathcal{G}$  & &  \\
  &   & are orthogonal & &  \\
  \hline
   OPFA & $\b{M}_s=\b{M}\left(\b{F},\b{d}^s\right)$, $\b{d}^s \in \mathcal{K}$ & Non-negative, & \tiny No & This work. \\
 & where $\b{F}$ is smooth &   sparse $\b{A}_s$ &   &   \\
 & and non-negative and &    &   &   \\
 & $\mathcal{K}$ enforces consistent  &     &   & \\
& precedence order &     &   & \\
  \hline
\end{tabular}
\end{center}
\end{table}

In this paper, we will restrict the
columns of $\b{M}_s$ to be translated versions of a common set of
factors, where these factors have onsets that occur in some relative
order that is consistent across all subjects. Our model differs from
previous shift-invariant models considered in \cite{lewicki1999coding},
\cite{blumensath2006sparse}, \cite{mailhe2008shift} in that it
restricts the possible shifts to those which preserve the relative
order of the factors among different subjects. We call the problem of
finding a decomposition (\ref{originalstructure}) under this
assumption the Order Preserving Factor Analysis (OPFA) problem.

The contributions of this paper are the following. First, we propose a
non-negatively constrained linear model that accounts for temporally
misaligned factors and order restrictions. Second, we give a computational
algorithm that allows us to fit this model in reasonable
time. Finally, we demonstrate that our methodology is able to
succesfully extract the principal features in a simulated dataset and in a real gene
expression dataset. In addition, we show that the application of OPFA
produces factors that can be used to significantly reduce the
variability in clustering of gene expression responses.

This paper is organized as follows. In Section \ref{sec:motivation} we
present the biological problem that motivates OPFA and introduce our
mathematical model. In Section \ref{sec:mathmodel}, we formulate the
non-convex optimization problem associated with the fitting of our
model and give a simple local optimization algorithm.  In Section
\ref{section:numresults} we apply our methodology to both synthetic
data and real gene expression data. Finally we conclude in Section
\ref{sec:conclusions}. For lack of space many technical details are
left out of our presentation but are available in the accompanying
technical report \cite{Puig&etal:TR2010}.

\section{Motivation: gene expression time-course data}
\label{sec:motivation}

In this section we motivate the OPFA mathematical model in the context
of gene expression time-course analysis. Temporal profiles of gene
expression often exhibit motifs that correspond to cascades of
up-regulation/down-regulation patterns. For example, in a study of a
person's host immune response after inoculation with a certain
pathogen, one would expect genes related to immune response to exhibit
consistent patterns of activation across pathogens, persons, and
environmental conditions.

\begin{figure}[t]
\centerline{\includegraphics[trim = 5mm 15mm 10mm 3mm, clip,width=270pt]{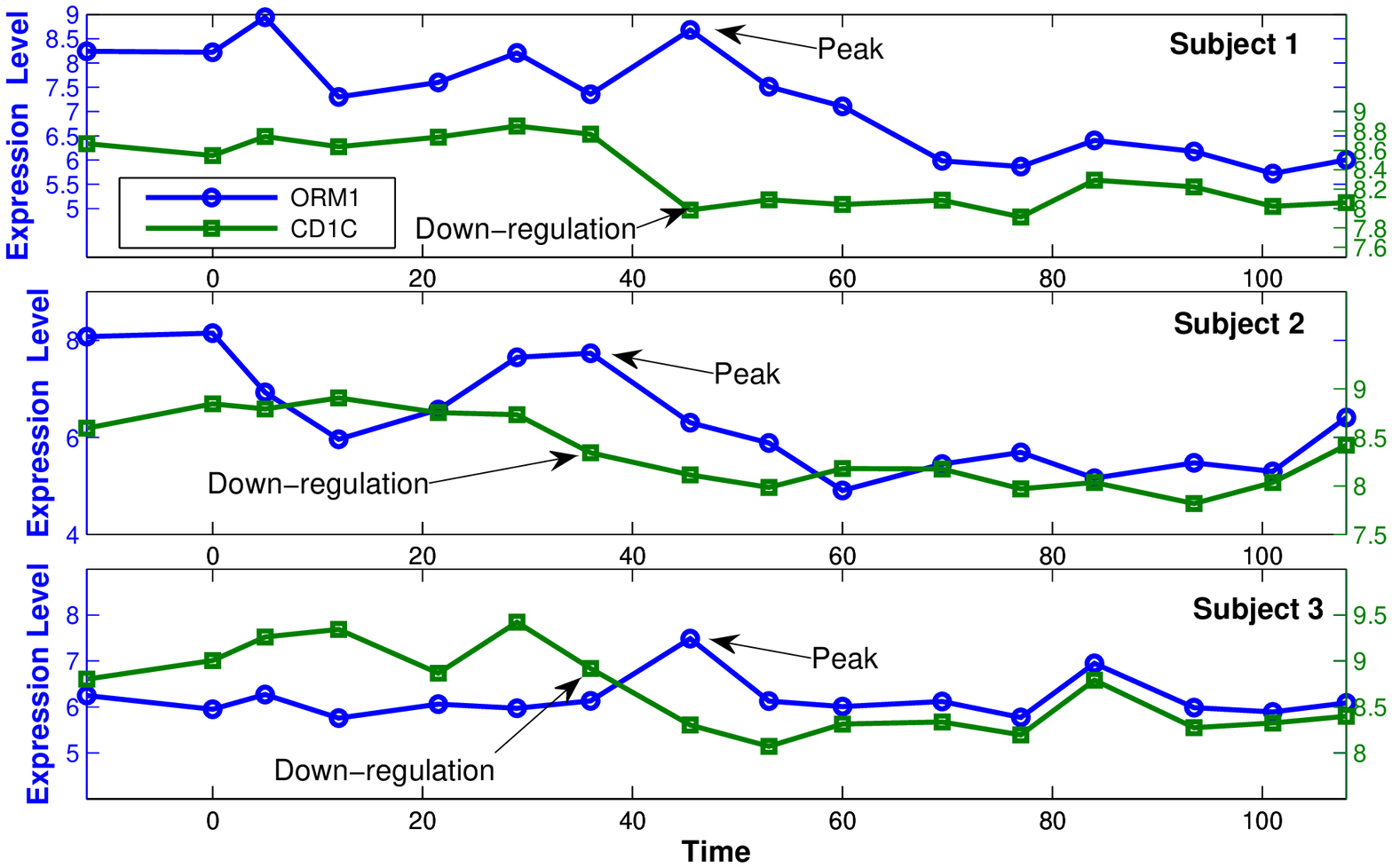}}
\caption{ Example of gene patterns with a consistent precedence-order across 3 subjects. 
The down-regulation motif of gene \textit{CD1C} precedes the peak motif of gene \textit{ORM1} across these three subjects.} 
\label{Fig:precedence}
\end{figure}

A simple approach to characterize the response patterns is to encode them as sequences of a few basic motifs such as 
(see, for instance, \cite{sacchi2007precedence}):
\begin{itemize}
\item \textit{Up-regulation}: Gene expression changes from low to high.
\item \textit{Down-regulation}: Gene expression changes from a high to a low level.
\item \textit{Steady}: Gene expression does not vary.
\end{itemize}
If gene expression is coherent over the population of several
individuals, e.g., in response to a common viral insult, the response
patterns can be expected to show some degree of consistency across
subjects. Human immune system response is a highly evolved system in
which several biological pathways are recruited and organized over
time. Some of these pathways will be composed of genes whose
expressions obey a precedence-ordering, e.g., virally induced
ribosomal protein production may precede toll-like receptor activation
and antigen presentation \cite{aderem2000toll}. This consistency
exists despite temporal misalignment: even though the order is
preserved, the specific timing of these events can vary across the
individuals. For instance, two different persons can have different
inflammatory response times, perhaps due to a slower immune system in
one of the subjects. This precedence-ordering of motifs in the
sequence of immune system response events is invariant to time shifts
that preserve the ordering. Thus if a motif in one gene precedes
another motif in another gene for a few subjects, we might expect the
same precedence relationship to hold for all other subjects. Figure
\ref{Fig:precedence} shows two genes from \cite{zaas2009gene} whose
motif precedence-order is conserved across 3 different
subjects. This conservation of order allows one to impose ordering
constraints on (\ref{originalstructure}) without actually knowing the
particular order or the particular factors that obey the
order-preserving property.

\begin{figure}[t]
\centerline{\includegraphics[trim =10mm 4mm 15mm 10mm, clip,width=250pt]{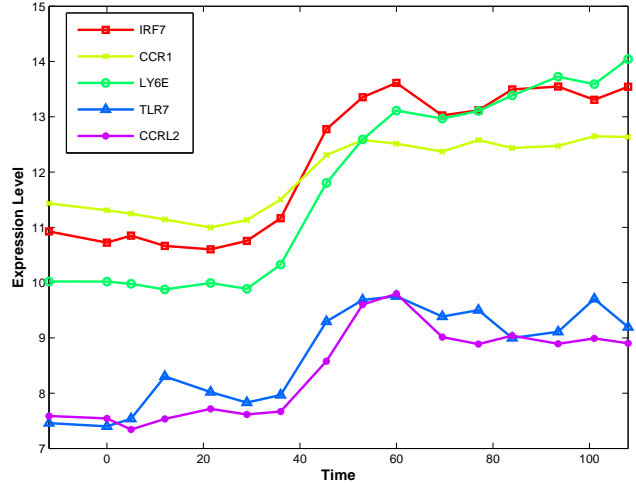}}
\caption{ Example of gene patterns exhibiting co-expression for a particular subject in the viral challenge study in \cite{zaas2009gene}.}
\label{Fig:coregulation}
\end{figure}

Often genes are co-regulated or co-expressed and have highly
correlated expression profiles. This can happen, for example, when the
genes belong to the same signaling pathway. Figure
\ref{Fig:coregulation} shows a set of different genes that exhibit a
similar expression pattern (up-regulation motif). The existence of high
correlation between large groups of genes allows one to impose a low rank
property on the factorization in (\ref{originalstructure}).

In summary, our OPFA model is based on the following assumptions:
\begin{itemize}
\item \textit{A1: Motif consistency across subjects}: Gene expression patterns have consistent (though not-necessarily
 time aligned) motifs across subjects undergoing a similar treatment.
\item \textit{A2: Motif sequence consistency across subjects}: If motif $X$ precedes motif $Y$ for subject $s$, 
the same precedence must hold for subject $t \neq s$.
\item \textit{A3: Motif consistency across groups of genes}: There are (not necessarily known) groups of genes 
that exhibit the same temporal expression patterns for a given subject.
\item \textit{A4: Gene Expression data is non-negative}: Gene expression on a microarray is measured as an
 abundance and standard normalization procedures, such as RMA \cite{irizarry2003exploration},
 preserve the non-negativity of this measurement.
\end{itemize}
A few microarray normalization software packages produce gene expression scores that do
not satisfy the non-negativity assumption A4. In such cases, the non-negativity  constraint in the  algorithm implementing
(\ref{JointProblem}) can be disabled. Note that in general, only a subset of genes may satisfy assumptions \textit{A1}-\textit{A3}. 
\section{OPFA mathematical model}
\label{sec:mathmodel}

In the OPFA model, each of the $S$ observations is
represented by a linear combination of \textit{temporally aligned}
factors. Each observation is of dimension $n \times p$, where $n$ is the number of time points and $p$ is the number of genes under consideration.  
Let $\b{F}$ be an $n \times f$
matrix whose columns are the $f$ common \textit{alignable} factors,
and let $\b{M}\left(\b{F},\b{d}\right)$ be a matrix valued function
that applies a circular shift to each column of $\b{F}$ according to the vector
of shift parameters $\b{d}$, as depicted in Figure \ref{Fig:scheme}. Then, we can refine model (\ref{originalstructure}) by restricting 
${\mathbf M}_s$ to have the form: 
\begin{eqnarray}
\label{factormodel}
\b{M}_{s}=\b{M}\left(\b{F},\b{d}^s\right).
\end{eqnarray}
where $\b{d}^s \in \left\{0,\cdots,d_{\max}\right\}^f$ and $d_{\max}\leq n$ is the maximum shift allowed in our model. 
This model is a generalization of a simpler one that restricts all factors to be aligned but with a common delay:
\begin{eqnarray}
\label{factormodelsimpler}
\b{M}_{s}=\b{U}_s \b{F},
\end{eqnarray}
where $\b{U}_s$ is a circular shift operator. 
Specifically, the fundamental characteristic of our model (\ref{factormodel}) is that each 
column can have a different delay, whereas (\ref{factormodelsimpler}) is 
a restriction of (\ref{factormodel}) with $d^s_i=d^s_j$ for all $s$ and all $i$, $j$.

 The circular shift is not restrictive. By embedding the observation
into a larger time window it can accommodate transient gene expression profiles in addition to periodic
ones, e.g., circadian rhythms \cite{Puig&etal:TR2010}. 
There are several ways to do this embedding. One way is to simply extrapolate the windowed, transient data to a 
larger number of time points $n_F=n+d_{\max}$. This is the strategy we follow in the numerical experiments of Section IV-B.

\begin{figure}[t]
\centerline{\subfigure{\includegraphics[trim = 0mm 10mm 0mm 10mm, clip,width=200pt]{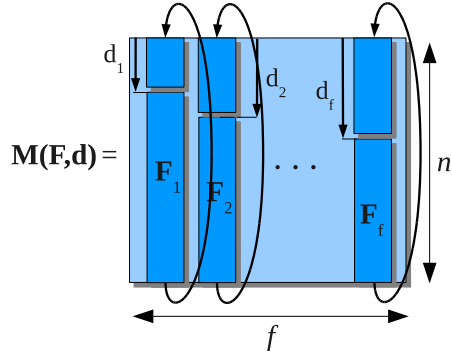}}}
\caption{Each subject's factor matrix $\b{M}_{s}$ is obtained by applying a circular shift to a common set of factors $\b{F}$ parameterized by a vector $\b{d}$.}
\label{Fig:scheme}
\end{figure}

This alignable factor model parameterizes each observation's intrinsic temporal dynamics through the $f$-dimensional vector $\b{d}^s$. The precedence-ordering constraint \textit{A2} is enforced by imposing
the condition
\begin{eqnarray}
\label{ordercnst}
d_{j_1}^{s_1} \leq d_{j_2}^{s_1} \Leftrightarrow d_{j_1}^{s_2} \leq d_{j_2}^{s_2} \mbox{ } \forall s_2 \neq s_1, 
\end{eqnarray}
that is, if factor $j_1$ precedes factor $j_2$ in subject $s_1$, then the same ordering will hold in all other subjects. Since the indexing of the factors is arbitrary, we can assume without loss of
generality that $d_i^s \leq d_{i+1}^s$ for all $i$ and all $s$. 
This characterization constrains each observation's delays $\b{d}^s$ independently, allowing for a computationally efficient algorithm for fitting model (\ref{factormodel}).

\subsection{Relationship to 3-way factor models.}

Our proposed OPFA framework is significantly
different from other factor analysis methods and these differences are
illustrated in the simulated performance comparisons below. However,
there are some similarities, especially to 3-way factor models \cite{kolda2009tensor}, \cite{comon2002tensor} that are worth pointing out. 

An $n$-th order tensor or $n$-way array is a data structure whose elements are indexed by an $n$-tuple of indices \cite{comon2002tensor}. $n$-way arrays can be seen as multidimensional generalizations of vectors and matrices: an $1$-way array is a vector and a $2$-way array is a matrix. Thus, we can view our observations $\b{X}_s$ as the slices of a third order tensor $\mathcal{X}$ of dimension $p \times n \times S$: $\b{X}_s= \mathcal{X}_{\cdot,\cdot,i}$. Tensor decompositions aim at extending the ideas of matrix (second order arrays) factorizations to higher order arrays \cite{kolda2009tensor}, \cite{comon2002tensor} and have found many applications in signal processing and elsewhere \cite{comon2002tensor}, \cite{kolda2009tensor},
\cite{Larsson2010HOSVD}, \cite{sidiropoulos2000blind},
\cite{de2000multilinear}. Since our data tensor is of order 3, we will only consider here 3-way decompositions, which typically take the following general form:
\begin{eqnarray}
\label{tuckerG}
\mathcal{X}_{i,j,k}= \sum_{p=1}^{P}\sum_{q=1}^{Q}\sum_{r=1}^{R} {\mathcal{G}}_{pqr} \b{F}_{ip}\b{B}_{jq}\b{C}_{kr} 
\end{eqnarray}
where $P$, $Q$, $R$ are the number of columns in each of the factor matrices $\b{F}$, $\b{B}$, $\b{C}$ and $\mathcal{G}$ is a $P \times Q \times R$ tensor. This class of decompositions is known as the Tucker model. When orthogonality is enforced among $\b{F}$, $\b{B}$, $\b{C}$ and different matrix slices of $\mathcal{G}$, one obtains the Higher Order SVD \cite{Larsson2010HOSVD}. When $\mathcal{G}$ is a superdiagonal tensor\footnote{$\mathcal{G}$ is superdiagonal tensor when $\mathcal{G}_{ijk}=0$ except for $i=j=k$.} and $P=Q=R$, this model amounts to the PARAFAC/Canonical Decomposition (CP) model \cite{carroll1970analysis}, \cite{sidiropoulos2000blind}, \cite{de2000multilinear}. The PARAFAC model is the closest to OPFA. Under this model, the slices of $\mathcal{X}_{i,j,k}$ can be written as:
\begin{eqnarray}
\label{parafac}
\b{X}^{\mbox{CP}}_s= \b{F}\diag\left(\b{C}_{\cdot,s}\right) \b{B}'.
\end{eqnarray}
This expression is to be compared with our OPFA model, which we state again here for convenience:
\begin{eqnarray}
\label{OPFAmodel}
\b{X}^{\mbox{OPFA}}_s=\b{M}\left(\b{F},\b{d}^s\right)\b{A}_s.
\end{eqnarray}
Essentially, (\ref{parafac}) shows that the PARAFAC decomposition
 is a special case of the OPFA model (\ref{OPFAmodel}) where the
 factors are fixed ($\b{M}_s=\b{F}$) and the scores only vary in
 magnitude across observations ($\b{A}_s=\diag\left(\b{C}_{\cdot,s}\right) \b{B}'$).
  This structure enhances uniqueness (under some conditions concerning the linear 
independence of the vectors in  $\b{F}$, $\b{B}$, $\b{C}$, see \cite{kolda2009tensor})
 but lacks the additional flexibility necessary to model possible translations in the 
columns of the factor matrix $\b{F}$. If $\b{d}^s=\b{0}$ for all $s$, then the OPFA (\ref{OPFAmodel}) 
and the Linear Factor  model (\ref{simplefamodel}) also coincide. The OPFA model can be 
therefore seen as an extension of the Linear Factor  and PARAFAC models where the
 factors are allowed to experiment order-preserving circular translations across different individuals.

\subsection{OPFA as an optimization problem}

OPFA tries to fit the model (\ref{originalstructure})-(\ref{ordercnst}) to the data $\left\{\b{X}_s\right\}_{s=1}^S$. For this purpose, we define the following penalized and
constrained least squares problem:
\begin{eqnarray}
\label{JointProblem}
\min &  \sum_{s=1}^S \l|\b{X}_s -  \b{M}\left(\b{F},\b{d}^s\right)\b{A}_s\r|_{F}^2+\lambda P_1\left(\b{A}_1,\cdots,\b{A}_S\right) + \beta P_2\left(\b{F}\right) \\ 
\mbox{ s.t. } & \left\{\b{d}^s\right\}_s \in {\mathcal{K}},\b{F} \in {\mathcal{F}} , \b{A}_s  \in {\mathcal{A}}_s \nonumber
\end{eqnarray}
where $\l|\cdot \r|_{F}$ is the Frobenius norm, $\lambda$ and $\beta$ are regularization parameters, and the set $\mathcal{K}$ constrains the delays $\b{d}^s$ to be order-preserving:
\begin{eqnarray}
\label{CosntrainCone}
{\mathcal{K}}=\left\{\b{d} \in \left\{0,\cdots,d_{\max}\right\}^{f} : {d}_{i+1} \geq {d}_{i}, \forall i \right\}. 
\end{eqnarray}
where $d_{\max} \leq n$. The other soft and hard constraints are briefly described as follows.

For the gene expression application we wish to extract factors
$\b{F}$ that are smooth over time and non-negative.
Smoothness will be captured by the constraint that $P_2(\b{F})$ is
small where $P_2(\b{F})$ is the squared total variation operator
\begin{eqnarray}
\label{limitedVariation}
 P_2\left(\b{F}\right) = \sum_{i=1}^f \l|\b{W} \b{F}_{\cdot, i} \r|_2^2
\end{eqnarray}
where $\b{W}$ is an appropriate weighting matrix and $\b{F}_{\cdot, i}$ denotes the $i$-th column of matrix $\b{F}$.
 From \textit{A4}, the data is non-negative and hence
non-negativity is enforced on $\b{F}$ and the loadings $\b{A}_s$ to
avoid masking of positive and negative valued factors whose overall
contribution sums to zero. To avoid numerical instability associated with
the scale invariance $\b{M}\b{A} = \frac{1}{\alpha}\b{M} \alpha \b{A}$
for any $\alpha >0$, we constrain the Frobenius norm of $\b{F}$. This
leads to the following constraint sets:
\begin{eqnarray}
\label{constrainSets}
{\mathcal{F}}=\left\{ \b{F} \in {\mathbb{R}}_{+}^{n \times f}: \l|\b{F}\r|_{F}\leq \delta \right\}\\
{\mathcal{A}}_s={\mathbb{R}}_{+}^{ f \times p}, s=1,\cdots,S \nonumber
\end{eqnarray}
The parameter $\delta$ above will be fixed to a positive value as its purpose is purely computational and has little practical impact.
Since the factors $\b{F}$ are common to all subjects, assumption \textit{A3} requires that the
number of columns of $\b{F}$ (and therefore, its rank) is small compared to the number of genes $p$. 
In order to enforce \textit{A1} we consider two different models. In the first model, which we shall name
{OPFA}, we constrain the columns of $\b{A}_{s}$ to be
sparse and the sparsity pattern to be consistent across different subjects. Notice that
\textit{A1} does not imply that the mixing weights $\b{A}_{s}$ are the same for all subjects as this 
would not accommodate magnitude variability across subjects. 
We also consider a more restrictive model where we constrain
 $\b{A}_{1}=\cdots=\b{A}_{S}=\b{A}$ with sparse $\b{A}$ and we call this model {OPFA-C}, 
the \textit{C} standing for the additional constraint that the subjects share the same sequence ${\mathbf A}$ of mixing weights.
 The {OPFA-C} model has a smaller number of parameters than {OPFA}, possibly at the expense of 
 introducing bias with respect to the unconstrained model.  A similar constraint has been succesfully adopted 
in \cite{jiafactorized} in a factor model for multi-view learning.

Similarly to the approach taken in \cite{mairal2010non} in the context of simultaneous sparse coding, 
the common sparsity pattern for {OPFA} is enforced by constraining $P_1\left(\b{A}_1,\cdots,\b{A}_S\right)$ to be small,
where $P_1$ is a mixed-norm group-Lasso type penalty function \cite{yuan2006model}.
 For each of the $p \times f$ score variables, we create a group containing its $S$ different values across subjects:
\begin{eqnarray}
\label{grouplasso}
P_1\left(\b{A}_1,\cdots,\b{A}_S\right)  = \sum_{i=1}^p\sum_{j=1}^f \|\left[\b{A}_1\right]_{j,i} \cdots \left[\b{A}_S \right]_{j,i}\|_2.
\end{eqnarray}
Table \ref{tab:models} summarizes the constraints of each of the models considered in this paper.


Following common practice in factor analysis, the non-convex problem (\ref{JointProblem}) is addressed using Block Coordinate Descent, displayed in
the figure labeled Algorithm \ref{FourierProblemAlgo}, which iteratively minimizes
(\ref{JointProblem}) with respect to the shift parameters
$\{\b{d}^s\}_{s=1}^S$, the scores $\{\b{A}_s\}_{s=1}^S$ and the factors $\b{F}$
while keeping the other variables fixed. This algorithm is guaranteed
to monotonically decrease the objective function at each iteration. Since both
the Frobenius norm and $P_1\left(\cdot\right)$,
$P_2\left(\cdot\right)$ are non-negative functions, this ensures that
the algorithm converges to a (possibly local) minima or a saddle point
of (\ref{JointProblem}).

{\begin{algorithm}[t!]
\caption{BCD algorithm for finding a local minima of (\ref{JointProblem}). 
$\mbox{OPFAObjective}\left(\b{F},\left\{ \b{A}_s\right\}_{s=1}^S ,\left\{ \b{d}^s\right\}_{s=1}^S\right)$ denotes
the objective function in (\ref{JointProblem})}
\label{FourierProblemAlgo}
\KwIn{Initial estimate of $\b{F}$ and $\left\{ \b{A}_s\right\}_{s=1}^S$, $\epsilon$, $\lambda$, $\beta$.}
\KwOut{$\b{F}$, $\left\{ \b{A}_s\right\}_{s=1}^S$, $\left\{ \b{d}^s\right\}_{s=1}^S$}
$c^0=\infty$\\
$c^{1}\leftarrow \mbox{OPFAObjective}\left(\b{F},\left\{ \b{A}_s\right\}_{s=1}^S ,\left\{ \b{d}^s\right\}_{s=1}^S\right)\nonumber $\\
$t=1$\\
\While{$c^{t-1}-c^{t} \geq \epsilon$}{
 $~~ \left\{ \b{d}^s\right\}_{s=1}^S\leftarrow\mbox{EstimateDelays}\left(\b{F},\left\{ \b{A}_s\right\}_{s=1}^S\right)$  \\
 $~~\left\{ \b{A}_s\right\}_{s=1}^S\leftarrow\mbox{EstimateScores}\left(\b{F},\left\{ \b{d}^s\right\}_{s=1}^S \right)$ \\
 $~~\b{F}\leftarrow\mbox{EstimateFactors}\left(\left\{ \b{A}_s\right\}_{s=1}^S,\left\{ \b{d}^s\right\}_{s=1}^S\right)$ \\
 $~~c^{t}\leftarrow \mbox{OPFAObjective}\left(\b{F},\left\{ \b{A}_s\right\}_{s=1}^S ,\left\{ \b{d}^s\right\}_{s=1}^S\right) $\\
 $~~t \leftarrow t+1$
}
\end{algorithm}
}

The subroutines $\mbox{EstimateFactors}$ and $\mbox{EstimateScores}$
solve the following penalized regression problems:
\begin{eqnarray}
\label{estimateFactors}
\min_{\b{F}} & \sum_{s=1}^S  \l|\b{X}_s -  \b{M}\left(\b{F},\b{d}^s\right)\b{A}_s\r|_{F}^2+  \beta \sum_{i=1}^f \l|\b{W} \b{F}_{\cdot, i} \r|_2^2 \\ 
\mbox{ s.t. } & \left\{
\begin{array}{c c}
 \l| \b{F} \r|^2_{\mathcal{F}} \leq \delta \nonumber & \\
 \b{F}_{i,j} \geq 0 & i=1,\cdots,n,\\
&  j=1,\cdots,f
\end{array}
\right.
\end{eqnarray}
and
\begin{eqnarray}
\label{estimateScores}
\min_{\left\{\b{A}_s\right\}_{s=1}^S} & \sum_{s=1}^S \l|\b{X}_s -  \b{M}\left(\b{F},\b{d}^s\right)\b{A}_s\r|_{F}^2+  \lambda \sum_{i=1}^p\sum_{j=1}^f \|\left[\b{A}_1\right]_{j,i} \cdots \left[\b{A}_S \right]_{j,i}\|_2 \\ 
\mbox{ s.t. } & \left\{
\begin{array}{l c}
 \left[\b{A}_s\right]_{j,i}\geq 0 & i=1,\cdots,n, \\
& j=1, \cdots,f,\\
&  s=1,\cdots,S
\end{array}
\right.
\end{eqnarray}
Notice that in OPFA-C, we also incorporate the constraint $\b{A}_{1}=\cdots=\b{A}_{S}$
in the optimization problem above. The
 first is a convex quadratic problem with a quadratic and a linear constraint
over a domain of dimension $f n$. In the applications considered here,
both $n$ and $f$ are small and hence this problem can be solved using any
standard convex optimization solver. $\mbox{EstimateScores}$ is trickier
because it involves a non-differentiable convex penalty and the dimension
of its domain is equal to\footnote{This refers to the OPFA model. In the OPFA-C model,
 the additional constraint $\b{A}_{1}=\cdots=\b{A}_{S}=\b{A}$ reduces the dimension
to $fp$.} $S f p$, where $p$ can be very large. In our
implementation, we use an efficient first-order method \cite{pustelnikconstrained}
designed for convex problems involving a quadratic term, a  non-smooth penalty 
and a separable constraint set. These procedures are described in more detail in
Appendix \ref{Appendix1} and
therefore we focus on the EstimateDelays subroutine. $\mbox{EstimateDelays}$ is a
discrete optimization that is solved using a branch-and-bound (BB)
approach \cite{lawler1966branch}.  In this approach a binary tree is created by recursively dividing the feasible set into subsets
(``branch'').  On each of the nodes of the tree lower and upper bounds
(``bound'') are computed. When a candidate subset is found whose upper
bound is less than the smallest lower bound of previously considered
subsets these latter subsets can be eliminated (``prune'') as candidate minimizers. Whenever a leaf (singleton subset) is obtained, the objective is evaluated at the corresponding point. If its value exceeds the current optimal value, the leaf is rejected as a candidate minimizer, otherwise the optimal value is updated and the leaf included in the list of candidate minimizers.
Details on the application of BB to OPFA are given below.

The subroutine $\mbox{EstimateDelays}$ solves $S$ uncoupled
problems of the form:
\begin{eqnarray}
\label{EstimateDelayproblem}
\min_{\bm{d} \in \mathcal{K}} \l|{\b{X}_s} -  \b{M}\left(\b{F},\b{d}\right)\b{A}_s \r|^2_{F},
\end{eqnarray}
where the set $\mathcal{K}$ is defined in (\ref{CosntrainCone}).  The
``branch'' part of the optimization is accomplished by recursive
splitting of the set $\mathcal{K}$ to form a binary tree. The recursion is
initialized by setting ${\mathcal{S}}_o=\left\{0,\cdots,d_{\max}\right\}^{f}$, 
${\mathcal{I}}_o=\left\{\b{d}\in {\mathcal{K}} \cap {\mathcal{S}}_o \right\}$. The splitting of
the set ${\mathcal{I}}_o$ into two subsets is done as follows
\begin{eqnarray}
\label{splifctn}
{\mathcal{I}}_1&=& 
\left\{\b{d}\in {\mathcal{K}} \cap {\mathcal{S}}_o : d_{\omega_1} \leq \gamma_1 \right\} \\
{\mathcal{I}}_2&=& 
 \left\{\b{d}\in {\mathcal{K}} \cap {\mathcal{S}}_o :  d_{\omega_1} > \gamma_1 \right\}, \nonumber
\end{eqnarray}
and we update ${\mathcal{S}}_1 = \left\{\b{d}\in {\mathcal{S}}_o :  d_{\omega_1} \leq \gamma_1 \right\}$, 
${\mathcal{S}}_2 = \left\{\b{d}\in {\mathcal{S}}_o :  d_{\omega_1} > \gamma_1 \right\}$. 
Here $\gamma_1$ is an integer $0 \leq \gamma_1 \leq d_{\max}$,
 and $\omega_1 \in \left\{1,\cdots,f\right\}$. ${\mathcal{I}}_1$ contains the elements $\b{d}\in{\mathcal{K}}$ 
whose $\omega_1$-th component is strictly larger than $\gamma_1$ and ${\mathcal{I}}_2$ contains 
the elements whose $\omega_1$-th component is smaller than $\gamma_1$.
%
The same kind of splitting procedure is then subsequently applied to
${\mathcal{I}}_1$, ${\mathcal{I}}_2$ and its resulting subsets.  After $k-1$
successive applications of this decomposition there will be $2^{k-1}$ subsets
and the $k$-th split will be :
\begin{eqnarray}
{\mathcal{I}}_{t}:=&\left\{\b{d} \in {\mathcal{K}} \cap {\mathcal{S}}_t \right\} \label{generalsubsetform}\\
{\mathcal{I}}_{t+1}:=&\left\{\b{d} \in {\mathcal{K}} \cap {\mathcal{S}}_{t+1} \right\}\nonumber \label{generalsubsetform2}
\end{eqnarray}  
where
\begin{eqnarray*}
{\mathcal{S}}_t = \left\{\b{d}\in {\mathcal{S}}_{\pi_k} :  d_{\omega_k} \leq \gamma_k \right\} \\
{\mathcal{S}}_{t+1} = \left\{\b{d}\in {\mathcal{S}}_{\pi_k} :  d_{\omega_k} > \gamma_k \right\}.
\end{eqnarray*} 
and $\pi_k \in \left\{1,\cdots, 2^{k-1}\right\} $ denotes the parent set of the two new sets $t$ and $t+1$, i.e.
$\mbox{pa}(t)=\pi_k$ and $\mbox{pa}(t+1)=\pi_k$. In our implementation the splitting coordinate $\omega_{k}$ is the one corresponding
to the coordinate in the set ${\mathcal{I}}_{\pi_k}$ with largest interval. The decision
point $\gamma_k$ is taken to be the middle point of this interval.

The ``bound'' part of the optimization is as
follows.  Denote $g\left(\b{d}\right)$ the objective function in
(\ref{EstimateDelayproblem}) and define its minimum over the set ${\mathcal{I}}_{t} \subset \mathcal{K} $:
\begin{eqnarray}
g_{\min}\left({\mathcal{I}}_t\right)=\min_{\b{d}\in {\mathcal{I}}_t} g\left(\b{d} \right).
\end{eqnarray}
A lower bound for this value can be obtained by relaxing the constraint $\b{d} \in \mathcal{K}$ in (\ref{generalsubsetform}):
\begin{eqnarray}
\label{LBRelaxation}
\min_{\b{d}\in {\mathcal{S}}_t } g\left(\b{d} \right) \leq g_{\min}\left({\mathcal{I}}_t\right) 
\end{eqnarray}
Letting $\b{X}_s=\b{X}_s^{\perp}+\b{X}_s^{\parallel}$ where $\b{X}_s^{\parallel}=\b{X}_s\b{A}_s^{\dag}\b{A}_s$ and $\b{X}_s^{\perp}=\b{X}_s\left(\b{I}-\b{A}_s^{\dag}\b{A}_s\right)$, we have:
\begin{eqnarray*}
\l|{\b{X}_s} -  \b{M}\left(\b{F},\b{d}\right)\b{A}_s\r|^2_{F}&= \l|\left({\b{X}_s}\b{A}_s^{\dag} -  \b{M}\left(\b{F},\b{d}\right)\right)\b{A}_s\r|^2_{F} \nonumber \\
 & + \l|{\b{X}_s}^{\perp} \r|^2_{F},  \nonumber
\end{eqnarray*}
where $\b{A}_s^{\dag}$ denotes the pseudoinverse of $\b{A}_s$. This leads to:
\begin{eqnarray}
\label{LBIneq}
\underline{\lambda}\left(\b{A}_s\b{A}_s^T\right)\l|{\b{X}_s}\b{A}_s^{\dag} -  \b{M}\left(\b{F},\b{d}\right) \r|^2_{F} +\l|{\b{X}_s}^{\perp} \r|^2_{F}\leq g\left(\b{d}\right), 
\end{eqnarray}
where $\underline{\lambda}\left(\b{A}_s\b{A}_s^T\right)$ denotes the smallest eigenvalue of the symmetric matrix $\b{A}_s\b{A}_s^T$. 
Combining the relaxation in (\ref{LBRelaxation}) with inequality  (\ref{LBIneq}), we obtain a lower bound on $g_{\min}\left({\mathcal{I}}_t\right)$:
\begin{eqnarray}
\label{LB}
\Phi_{lb}\left({\mathcal{I}}_t\right)&= {\min}_{\b{d}\in {\mathcal{S}}_t} \underline{\lambda}\left(\b{A}_s\b{A}_s^T\right)\l|{\b{X}_s}\b{A}_s^{\dag} -  \b{M}\left(\b{F},\b{d}\right) \r|_{F}^2 \nonumber\\ 
& +\l|{\b{X}_s}^{\perp} \r|^2_{F} \nonumber \\
& \leq g_{\min}\left({\mathcal{I}}_t\right),\label{eq:lowerbnd}
\end{eqnarray}
which can be evaluated by performing $f$ \textit{decoupled}
discrete grid searches. At the $k$-th
step, the splitting node $\pi_k$ will be chosen as the one with smallest $\Phi_{lb}\left({\mathcal{I}}_t\right)$.
Finally, this lower bound is complemented by the upper bound
\begin{eqnarray}
\label{UB}
g_{\min}\left({\mathcal{I}}_t\right) \leq \Phi_{ub}\left({\mathcal{I}}_t\right)=  g\left(\bm{d} \right) \mbox{ for } 
 \forall \bm{d} \in {\mathcal{I}}_t.
\end{eqnarray}
These bounds enable the branch-and-bound optimization of 
(\ref{EstimateDelayproblem}).

\subsection{Selection of the tuning parameters $f$, $\lambda$ and $\beta$}
\label{sec:CV}

From (\ref{JointProblem}), it is clear that the OPFA factorization  depends on
the choice of $f$, $\lambda$ and $\beta$. This is a paramount problem
in unsupervised learning, and several heuristic approaches have been devised
for simpler factorization models \cite{owen2009bi,wold1978cross,witten2009penalized}.
These approaches are based on training the factorization model on a subset of the elements
of the data matrix  (training set) to subsequently validate it on the excluded
elements (test set). 

The variational characterization of the OPFA decomposition allows
for the presence of missing variables, i.e. missing elements
in the observed matrices $\left\{\b{X}_s\right\}_{s=1}^S$. In such case,
the Least Squares fitting term in (\ref{JointProblem}) is 
only applied to the observed set of indices\footnote{See the Appendix \ref{Appendix1} and \ref{Appendix2} for the extension
of the EstimateFactors, EstimateScores and Estimatedelays procedures to the case where there exist
 missing observations.}.
 We will hence follow the approach in \cite{witten2009penalized}
and train the OPFA model over a fraction $1-\delta$ of the entries
in the observations $\b{X}_s$. Let $\Omega_s$ denote
the set of $\delta \left( n \times p \right)$ excluded entries for the $s$-th observation.
These entries will constitute our test set, and thus our Cross-Validation error measure 
is:
\begin{eqnarray*}
 \mbox{CV}\left(f,\lambda,\beta\right) =\frac{1}{S} \sum_{s=1}^S \l|\left[\b{X}_s - \b{M}\left(\hat{\b{F}},\hat{\b{d}}^s \right)\hat{\b{A}}_s \right]_{\Omega_s}\r|_{F}^2
\end{eqnarray*}
where $\hat{\b{F}},\left\{\hat{\b{d}}^s\right\}_{s=1}^S,\left\{\hat{\b{A}}_s\right\}_{s=1}^S$
are the OPFA estimates obtained on the training set excluding the entries in $\left\{\Omega_s\right\}_{s=1}^S$,
for a given choice of $f$, $\lambda$, and $\beta$.

\section{Numerical results}
\label{section:numresults}
\subsection{Synthetic data: Periodic model}
\label{section:numresultsSynth}

\begin{figure}[t]
\centerline{\includegraphics[trim = 10mm 0mm 10mm 0mm, clip,width=260pt]{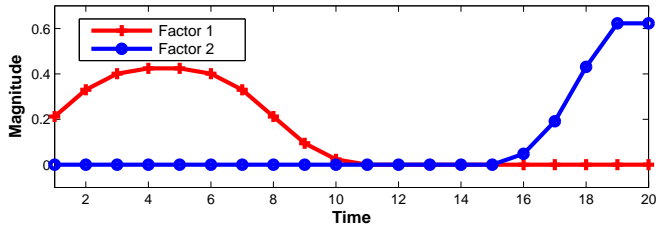}}
\caption{Dictionary used to generated the 2-factor synthetic data of Section \ref{section:numresults}.} 
\label{fig:ExDictionary}
\end{figure}

First we evaluate the performance of the OPFA algorithm for a 
periodic model observed in additive Gaussian white noise:
\begin{eqnarray}
\label{gendatamodel}
\b{X}_s=\b{M}\left(\b{F},\b{d}^s\right)\b{A}_s + \epsilon_s &  s=1, \ldots, S.
\end{eqnarray}
Here $\epsilon_s \sim {\mathcal{N}}_{n \times
  p}\left(\b{0},\sigma_\epsilon^2 \b{I} \right)$, $\b{d}^{s}=
\mbox{sort}\left( \b{t}^s\right)$ where $\sigma_{\epsilon}^2$ is the variance of $\epsilon_s$ and $\b{t}^s\sim
{\mathcal{U}}\left(0,\sqrt{12\sigma_d^2+1}\right)$ are i.i.d. The $f=2$ columns of
$\b{F}$ are non-random smooth signals from the predefined dictionary
shown in Figure \ref{fig:ExDictionary}. The scores $\b{A}_s$ are
generated according to a consistent sparsity pattern across all subjects
and its non zero elements are i.i.d. normal truncated to the non-negative orthant.

\begin{table}[t]
\caption{\small \label{Table:review} Models considered in Section IV-A. } 
\label{tab:models}
\vspace{-10pt}
\begin{center}
\footnotesize{
\begin{tabular}{|c|c|c|}
  \hline
   \textbf{Model} & \textbf{$\b{M}_s$} & \textbf{$\b{A}_s$} \\
  \hline
   OPFA & $\b{M}_s=\b{M}\left(\b{F},\b{d}^s\right)$ & Non-negative  \\ 
 &  $\b{d}^s \in \mathcal{K}$, $\b{F}$ smooth &  sparse $\b{A}_s$\\ 
 & and non-negative &  \\
  \hline
   OPFA-C & $\b{M}_s=\b{M}\left(\b{F},\b{d}^s\right)$ & Non-negative  \\ 
 &  $\b{d}^s \in \mathcal{K}$, $\b{F}$ smooth &  sparse \\ 
 & and non-negative &  $\b{A}_1=\cdots=\b{A}_S$\\
  \hline
   SFA & $\b{M}_s=\b{M}\left(\b{F},\b{d}^s\right)$, & Non-negative \\
 &   $\b{d}^s=\b{0}$ ,$\b{F}$ smooth  &   sparse $\b{A}_s$   \\
 & and non-negative& \\ 
  \hline
\end{tabular}}
\end{center}
\vspace{-10pt}
\end{table}

Here the number of subjects is $S=10$, the number of variables is
$p=100$, and the number of time points is $n=20$.  In these experiments experiments we choose 
to initialize the factors $\b{F}$ with temporal profiles obtained by hierarchical 
clustering of the data. Hierarchical clustering \cite{hastie2005elements} is a standard unsupervised learning technique that
groups the $p$ variables into increasingly finer partitions according to the normalized euclidean distance of their temporal profiles. 
The average expression patterns of the clusters found are used as initial estimates for $\b{F}$.
The loadings $\left\{ \b{A}_s\right\}_{s=1}^S$ are initialized by regressing the obtained factors onto the data.

We compare OPFA and OPFA-C to a standard Sparse Factor Analysis (SFA) solution, obtained by
imposing $d_{\max}=0$ in the original OPFA model. Table \ref{tab:models} summarizes 
the characteristics of the three models considered in the simulations. 
We fix $f=2$ and choose the tuning parameters $\left(\lambda,\beta\right)$ using
the Cross-Validation procedure of Section \ref{sec:CV} with a $5 \times 3$ grid
and $\delta=.1$. 

In these experiments, we consider two measures of performance, the Mean Square Error (MSE) with respect to the generated data:
\begin{eqnarray*}
MSE:=\frac{1}{S}\sum_{s=1}^{S}E\l|\b{D}_s-\hat{\b{D}}_s\r|_F^2,
\end{eqnarray*}
where $E$ is the expectation operator, $\b{D}_s=\b{M}\left(\b{F},\b{d}^s\right)\b{A}_s$ is the
generated noiseless data and
$\hat{\b{D}}_s=\b{M}\left(\hat{\b{F}},\hat{\b{d}}^s\right)\hat{\b{A}}_s$
is the estimated data, and the Distance to the True Factors (DTF),
defined as:
\begin{eqnarray*}
DTF:=1-\frac{1}{f}\sum_{i=1}^{f}E\frac{\b{F}_{\cdot,i}^T\hat{\b{F}}_{\cdot,i}}{\l|\b{F}_{\cdot,i}\r|_2\l|\hat{\b{F}}_{\cdot,i}\r|_2},
\end{eqnarray*}
where $\b{F}$, $\hat{\b{F}}$ are the generated and the estimated factor matrices, respectively.

\begin{figure}[htb]
\begin{minipage}[b]{1\linewidth}
  \centering
 \centerline{\includegraphics[trim = 0mm 0mm 0mm 0mm, clip,width=200pt]{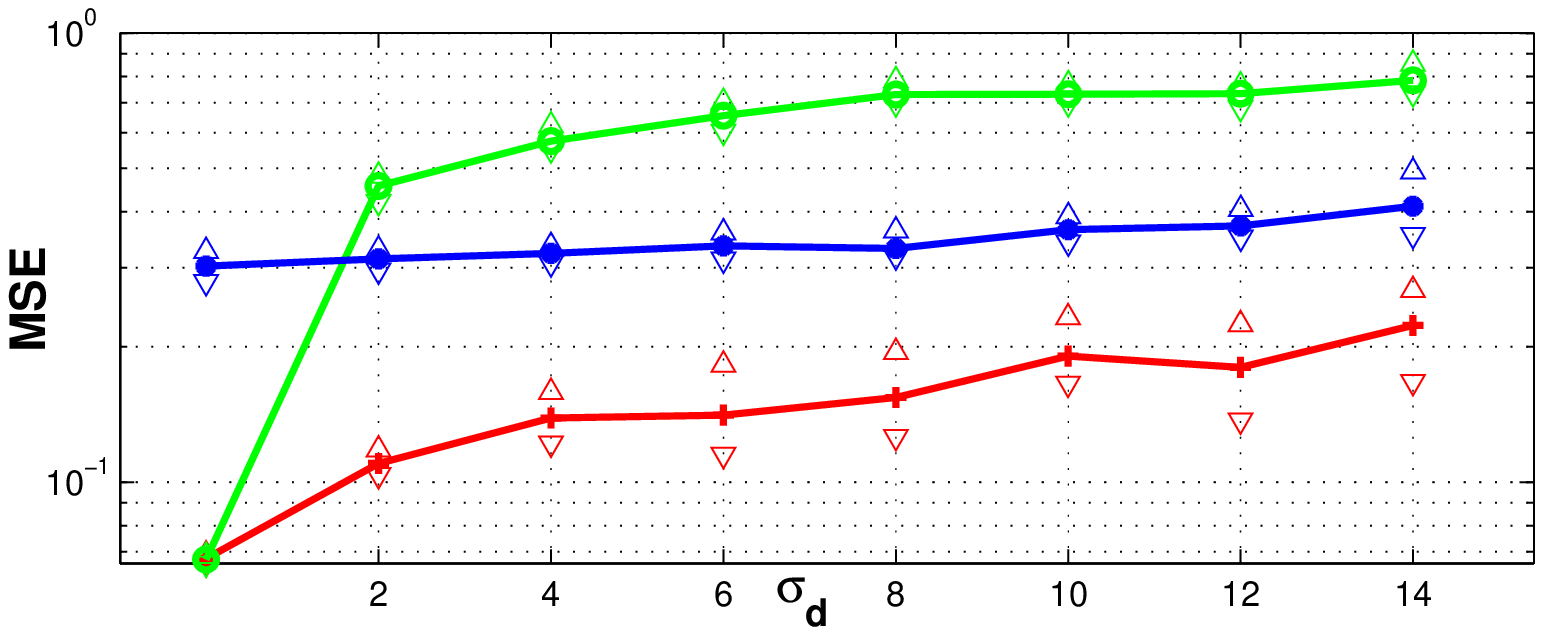}}
\end{minipage}
\vfill
\begin{minipage}[b]{1\linewidth}
  \centering
 \centerline{\includegraphics[trim = 0mm 0mm 0mm 0mm, clip,width=200pt]{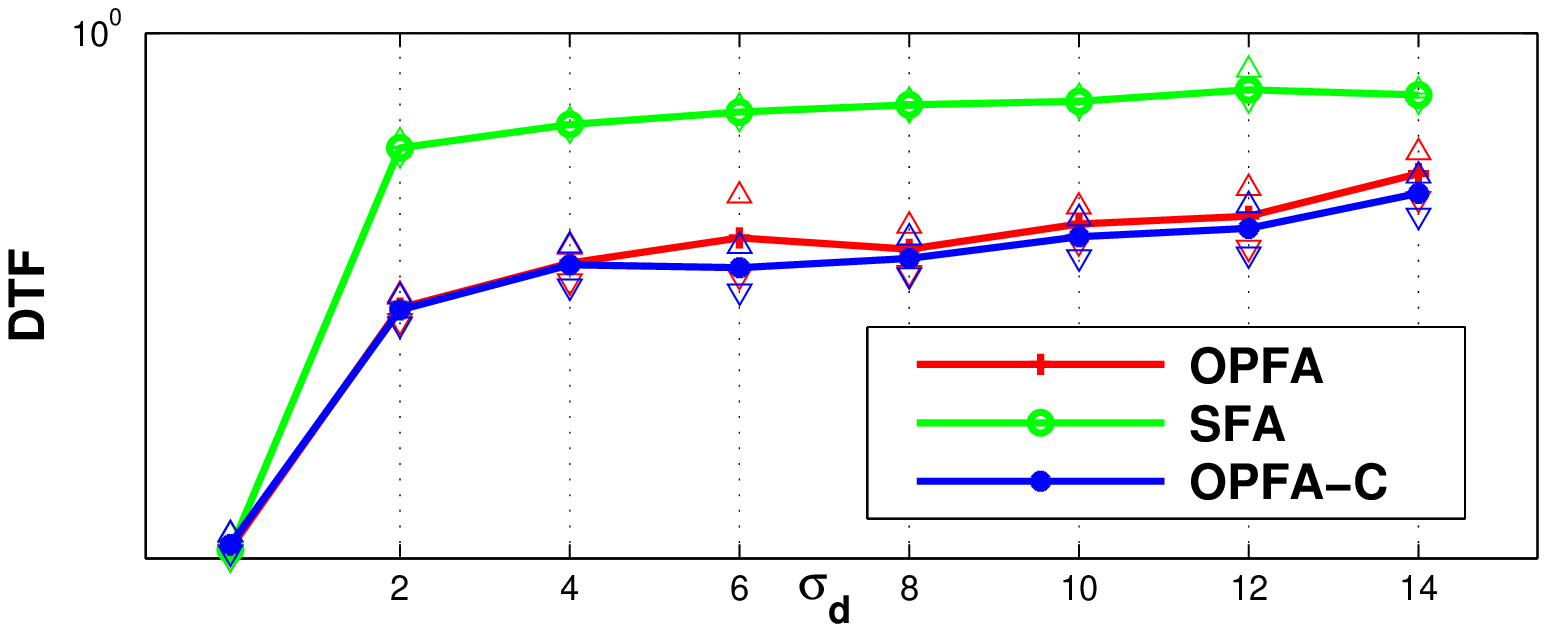}}

\end{minipage}
\caption{MSE (top) and DTF (bottom) as a function of delay variance
  $\sigma_d^2$ for OPFA and Sparse Factor Analysis (SFA). These
  curves are plotted with  95\% confidence intervals.  For $\sigma_d^2 > 0$, OPFA outperforms SFA both in MSE and DTF, maintaining its advantage as $\sigma_d$ increases. For large $\sigma_d$, OPFA-C outperforms the other two.}
\label{fig:Exp1}
\end{figure}

\begin{figure}[t]
\begin{minipage}[b]{1\linewidth}
  \centering
 \centerline{\includegraphics[trim = 0mm 0mm 0mm 0mm, clip,width=200pt]{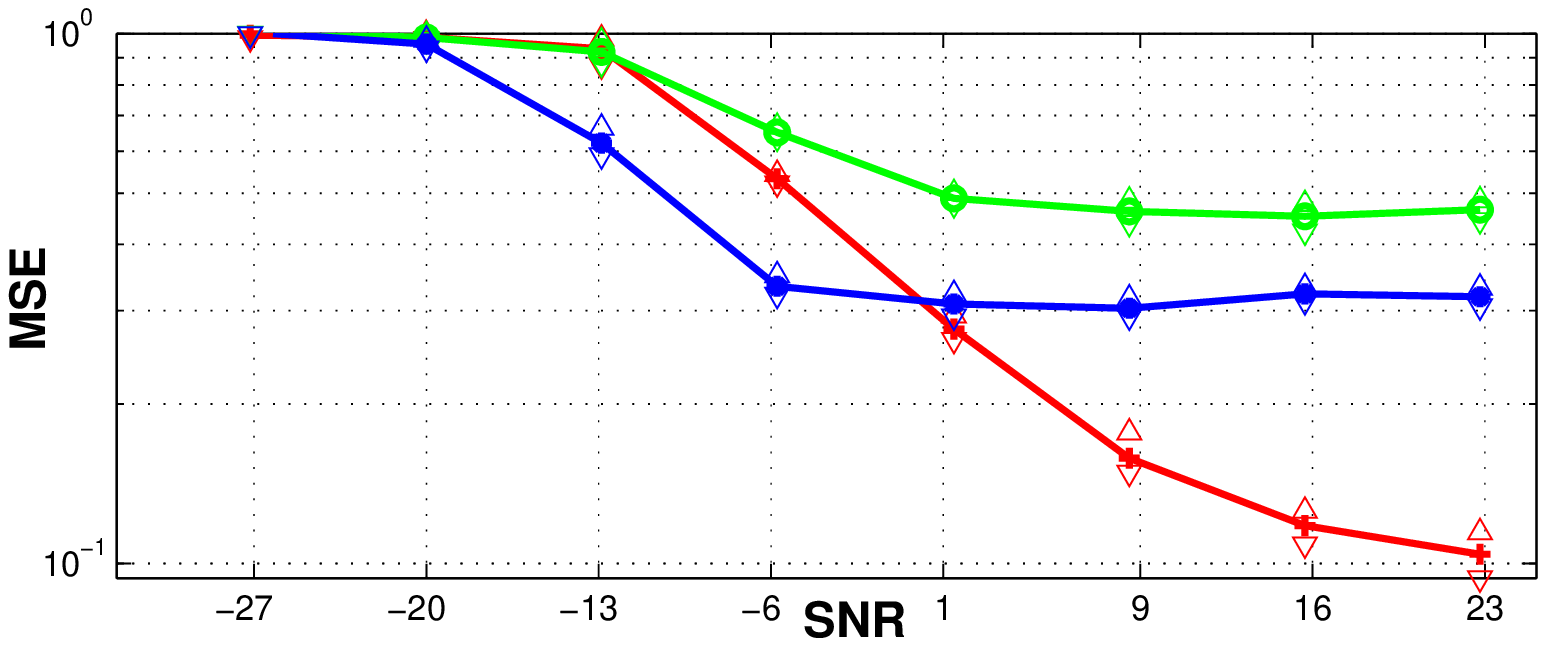}}

\end{minipage}
\vfill
\begin{minipage}[b]{1\linewidth}
  \centering
 \centerline{\includegraphics[trim = 0mm 0mm 0mm 0mm, clip,width=200pt]{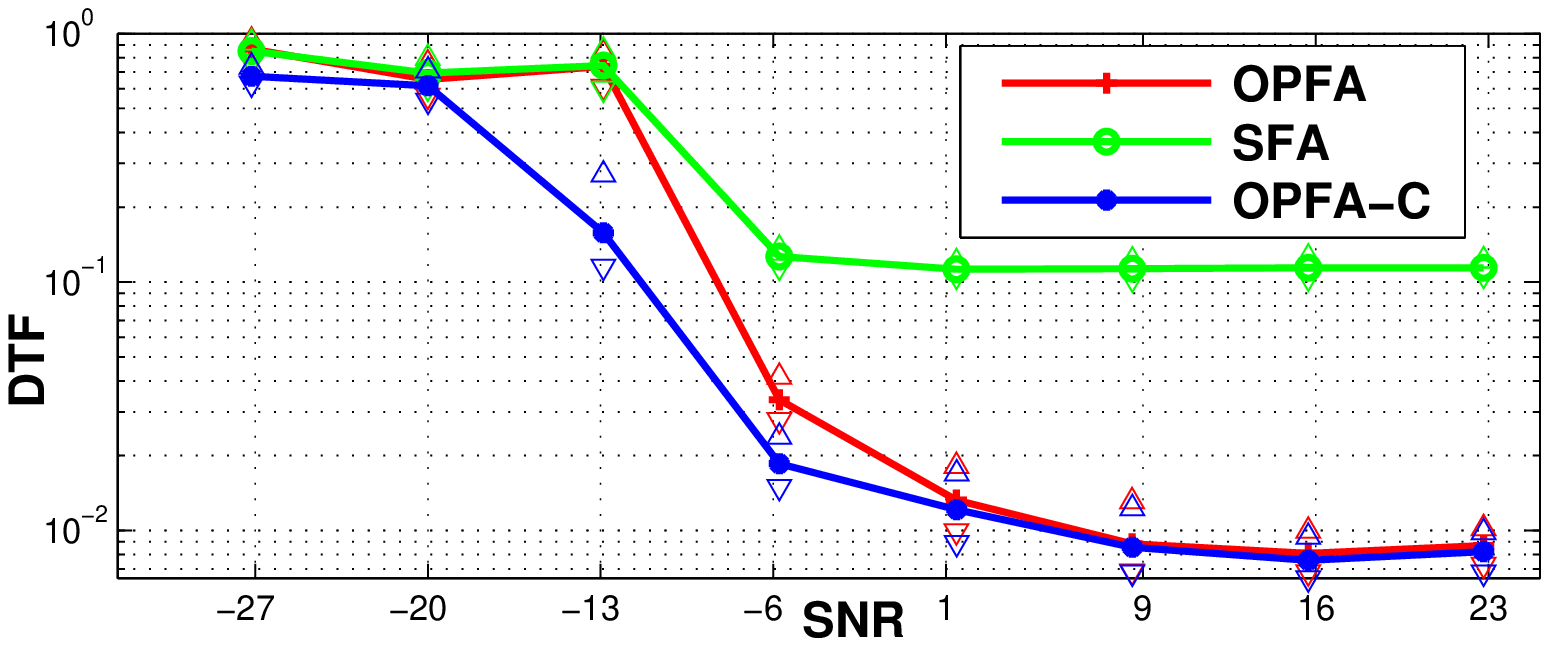}}

\end{minipage}
\caption{Same as Figure \ref{fig:Exp1} except that the performance curves are plotted with respect to SNR for fixed $\sigma_d^2=5$.} 
\label{fig:Exp2}
\end{figure}

Figure \ref{fig:Exp1} shows the estimated MSE and DTF performance curves as a
function of the delay variance $\sigma_d^2$ for fixed SNR$=15$dB 
(which is defined as 
$SNR=10\log\left(\frac{1}{S}\sum_s\frac{E\left( \l|\b{M}\left(\b{F},\b{d}^s\right)\b{A}_s\r|^2_{F}\right)}{np\sigma_{\epsilon}^2} \right)$).
 OPFA and OPFA-C perform at least as well as SFA for zero delay ($\sigma_d=0$) and
significantly better for $\sigma_d>0$ in terms of DTF. OPFA-C outperforms OPFA for high delay variances  $\sigma_d^2$ at the price of a larger
MSE due to the bias introduced by the constraint $\b{A}_1 = \cdots =\b{A}_S$. In Figure
\ref{fig:Exp2} the performance curves are plotted as a function of
SNR, for fixed
$\sigma_d^2=5$. Note that OPFA and OPFA-C outperform SFA in terms of DTF and that OPFA is better than the others in terms of MSE for SNR$>0$db. Again, OPFA-C 
shows increased robustness to noise in terms of DTF.

We also performed simulations to demonstrate the value of imposing the order-preserving constraint in (\ref{EstimateDelayproblem}). 
This was accomplished by comparing OPFA to a version of OPFA for which the constraints in (\ref{EstimateDelayproblem})
 are not enforced. Data was generated according to the model (\ref{gendatamodel}) with  $S=4$, $n=20$, $f=2$, and $\sigma_d^2=5$. 
The results of our simulations (not shown)  were that, while  the order-preserving constraints never degrade OPFA performance,
  the constraints  improve performance when the SNR is small (below  3dB for this example). 

Finally, we conclude this sub-section by studying the sensitivity of the final OPFA estimates 
with respect to the initialization choice. To this end, we initialize the OPFA algorithm with
the correct model perturbed with a random gaussian vector of increasing variance. We analyze the performance of the estimates in terms of
MSE and DTF as a function of the norm of the model perturbation relative to the norm of the noiseless data, which we denote by $\rho$. Notice that 
larger $\rho$ corresponds to increasingly random initialization. The results in Table \ref{tab:sensitivity} show that the MSE and DTF of the OPFA estimates
are very similar for a large range of values of $\rho$, and therefore are robust to the initialization. 

\begin{table}[t]
\caption{\label{tab:sensitivity} \small Sensitivity of the OPFA estimates to the initialization choice
 with respect to the relative norm of the perturbation ($\rho$).} 
\vspace{-10pt}
{\scriptsize 
\begin{center}
\begin{tabular}{|c|c|c|c|}
\hline
 & \multicolumn{3}{|c|}{DTF [ mean (standard deviation) ] $\times 10^{-3}$ }  \\
  \hline
 SNR    & $\rho=0.002$ &   $\rho=1.08$ & $\rho=53.94$ \\
  \hline
$22.8$ & $0.0$ ($0.0$) &  $3.4$ ($9.4 $) & $1.9$ ($3.2 $) \\
  \hline
$-2$&   $1.3$ ($0.5$)  & $1  $ ($9.4$) & $1.25$ ($1.5$)\\
  \hline
$-27.1$ & $46$ ($20$) &  $58$ ($17$) & $63$ ($8$)\\
  \hline
\end{tabular}



\begin{tabular}{|c|c|c|c|}
\hline
 & \multicolumn{3}{|c|}{MSE [ mean (standard deviation $\times 10^{-3}$) ]}  \\
  \hline
 SNR    & $\rho=0.002$ &   $\rho=1.08$ & $\rho=53.94$ \\
  \hline
$22.8$ & $0.02$ ($1.5 $) &  $0.05 $ ($69$) & $0.11$ ($99$) \\
  \hline
$-2$&  $0.35$ ($7.9 $) & $0.36$($22 $)&$0.38$ ($32 $)\\
  \hline
$-27.1$ & $0.96$ ($19 $) &  $ 0.99$ ($18 $) & $1.00$ ($24$)\\
  \hline
\end{tabular}
\end{center}
\vspace{-10pt}
}


\end{table}

\subsection{Experimental data: Predictive Health and Disease (PHD)}
\label{section:phddata}
The PHD data set was collected as part of a viral challenge study that is
described in \cite{zaas2009gene}. In this study 20 human subjects
were inoculated with live H3N2 virus and Genechip mRNA gene expression
in peripheral blood of each subject was measured over 16 time points. The raw Genechip array data was 
pre-processed using robust multi-array analysis \cite{irizarry2003exploration} with quantile normalization \cite{bolstad2003comparison}.
In this section we show results for the constrained OPFA model (OPFA-C). While not shown here, 
we have observed that  OPFA-C gives very similar results to unconstrained OPFA but with 
reduced computation time. 

Specifically, we use OPFA-C to perform the following tasks:
\begin{enumerate}
\item \textit{Subject Alignment:} Determine the alignment of the factors to fit each subject's 
response, therefore revealing each subject's intrinsic response delays.
\item \textit{Gene Clustering:} Discover groups of genes with common expression signature by clustering 
in the low-dimensional space spanned by the aligned factors. Since we are using the OPFA-C model, the projection of each subject's data on this
 lower dimensional space is given by the scores  $\b{A} := \b{A}_1=\cdots =\b{A}_S$. Genes with similar scores 
will have similar expression signatures.
\item \textit{Symptomatic Gene Signature discovery:} Using the gene clusters obtained in step 2 we construct 
temporal signatures common to subjects who became sick. 
\end{enumerate}

The data was normalized by dividing each element of each data matrix by the sum of the elements in its column. Since the data is non-negative valued, 
this will ensure that the mixing weights of different subjects are within the same order of magnitude, which is necessary to 
respect the assumption that $\b{A}_{1}=\cdots=\b{A}_{S}$ in OPFA-C. In order to select a subset of strongly varying genes,
we applied one-way Analysis-Of-Variance \cite{neter1996applied}
to test for the equality of the mean of each gene at 4 different groups of time points, and selected the first $p=300$ genes ranked according to
the resulting F-statistic. To these gene trajectories we applied OPFA-C to the $S=9$ symptomatic subjects in
the study. In this context, the columns in $\b{F}$ are the set of signals emitted by the common
immune system response and the vector $\b{d}^s$ parameterizes each subject's
characteristic onset times for the factors contained in the columns of
$\b{F}$. To avoid wrap-around effects, we worked with a factor model of dimension
$n=24$ in the temporal axis.

The OPFA-C algorithm was run with 4
random initializations and retained the solution yielding the minimum
of the objective function (6).  For each $f=1,\cdots,5$ (number of factors), we estimated
the tuning parameters $(\lambda, \beta)$ following the Cross-Validation approach described
in \ref{sec:CV} over a $10 \times 3$ grid. The resulting results, shown
in Table \ref{tab:error} resulted in selecting $\beta=1 \times 10^{-8}$, $\lambda= 1 \times 10^{-8}$
 and $f=3$. The choice of three factors is also consistent with an expectation that the principal
gene trajectories over the period of time studied are a linear combination of increasing, decreasing
or constant expression patterns \cite{zaas2009gene}.

\begin{table}[t]
\caption{\label{Table:relerr} \small Cross Validation Results for Section \ref{section:phddata}.} 
\label{tab:error}
\vspace{-10pt}
\begin{center}
\footnotesize{
\centering \begin{tabular}{|c |c|c|c|c|c|}
  \hline
   & $f=1$& $f=2$ &$\mathbf{f=3}$ & $f=4$ & $f=5$ \\
  \hline
   $\min \mbox{CV}\left(f,\lambda,\beta\right)$ &  20.25 & 13.66 & \textbf{12.66} & 12.75 & 12.72 \\
  \hline 
   Relative residual&  7.2 & 4.8 & \textbf{4.5} & 4.5 & 4.4 \\
    error ($\times 10^{-3}$)&   &  &  & &  \\
  \hline 
   $\hat{\lambda}$($\times 10^{-8}$) &  5.99 & 1 & \textbf{1} & 1 & 35.9 \\
  \hline 
   $\hat{\beta}$ ($\times 10^{-6}$)&  3.16 & 3.16 & \textbf{0.01} & 0.01 & 100 \\
  \hline 
\end{tabular}}
\end{center}
\vspace{-10pt}
\end{table}

\begin{figure}[t]
\centerline{\includegraphics[trim = 0mm 0mm 5mm 0mm, clip,height=200pt]{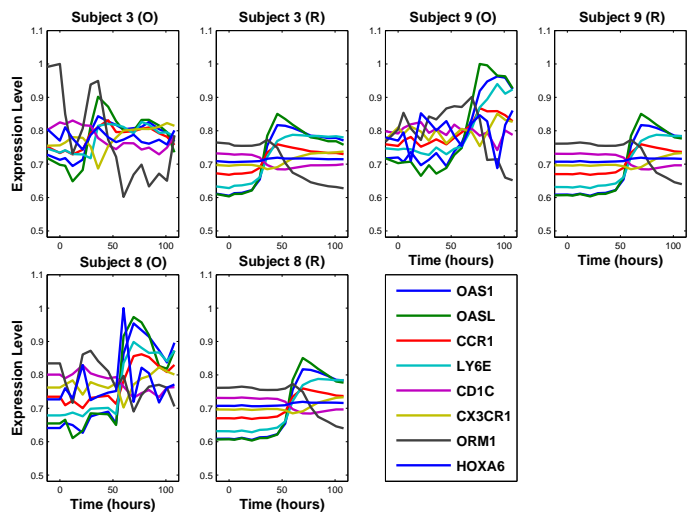}}
\caption{Comparison of observed (O) and fitted responses (R) for three of the
  subjects and a subset of genes in the PHD data set. Gene expression
  profiles for all subjects were reconstructed with a relative residual error
  below 10\%. The trajectories are smoothed while respecting each subject's intrinsic delay.}
\label{Fig:comparisonrealfittedSx}
\end{figure}

\begin{figure}[t]
\centerline{\includegraphics[trim = 0mm 0mm 10mm 0mm, clip,height=200pt]{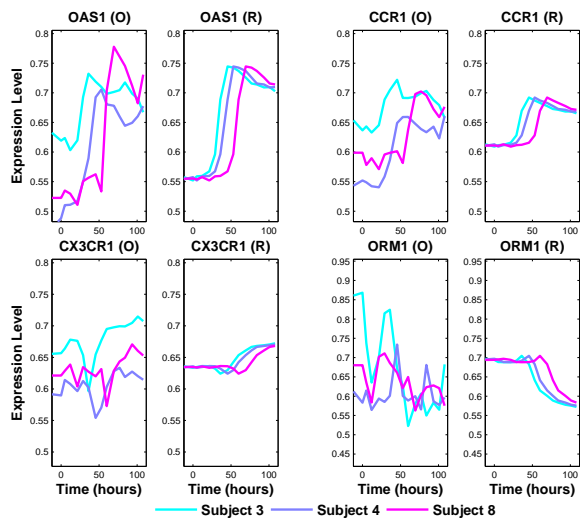}}
\caption{Comparison of observed (O) and fitted responses (R) for four
  genes (\textit{OAS1}, \textit{CCR1}, \textit{CX3CR1}, \textit{ORM1}) showing up-regulation and down-regulation motifs
  and three subjects in the PHD dataset. The gene trajectories have been smoothed while conserving their temporal pattern and their precedence-order. 
The OPFA-C model revealed that
  \textit{OAS1} up-regulation occurs consistently after \textit{ORM1} down-regulation among all
  subjects. }
\label{Fig:comparisonrealfittedSxGeneWise}
\end{figure}

To illustrate the goodness-of-fit of our model, we plot in Figure
\ref{Fig:comparisonrealfittedSx} the observed gene expression patterns
of 13 strongly varying genes and compare them to the OPFA-C fitted response
for three of the subjects, together with the relative residual
error. The average relative residual error is below $10\%$ and the plots
demonstrate the agreement between the observed and the fitted
patterns. Figure \ref{Fig:comparisonrealfittedSxGeneWise} shows the trajectories for each subject
for four genes having different regulation motifs: up-regulation and down-regulation. It is clear that the gene
trajectories have been smoothed while conserving their temporal pattern and their precedence-order,
 e.g. the up-regulation of gene \textit{OAS1} consistently follows the down-regulation of gene \textit{ORM1}.

\begin{figure*}[t]
\centerline{\subfigure{\includegraphics[trim =0mm 22mm 0mm 0mm, clip,width=350pt]{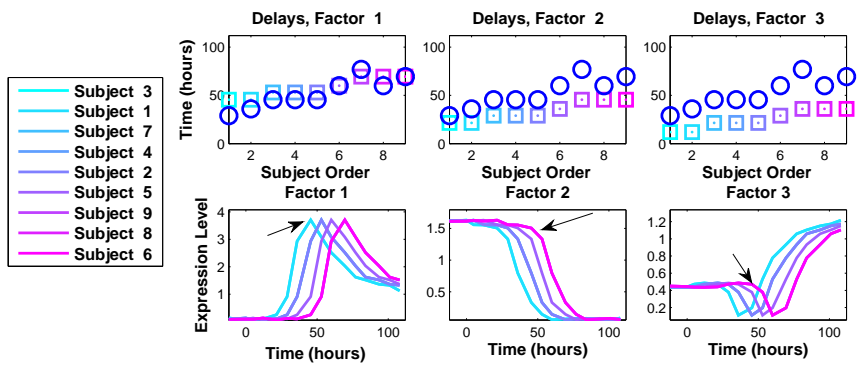}}}

\caption{Top plots: Motif onset time for each factor (\Square) and peak
  symptom time reported by expert clinicians (O). Bottom plots: Aligned
  factors for each subject. Factor 1 and 3 can be interpreted as up-regulation
  motifs and factor 2 is a strong down-regulation pattern. The arrows
show each factor's motif onset time.}
\label{Fig:delayssubjects}
\end{figure*}

In Figure \ref{Fig:delayssubjects} we show the 3 factors along with the
factor delays and factor loading discovered by OPFA-C.  The three
factors, shown in the three bottom panels of the figure, exhibit
features of three different motifs: factor 1 and
factor 3 correspond to up-regulation motifs occurring at different times; and factor 2 is a strong
down-regulation motif. The three top panels show the onset times of
each motif as compared to the clinically determined peak symptom onset
time. Note, for example, that the  strong up-regulation pattern of the first
factor coincides closely with the onset peak time. Genes strongly associated to this factor have
been closely associated to acute anti-viral and inflammatory host
response \cite{zaas2009gene}. Interestingly, the down-regulation
motifs associated with factor 2 consistently precedes this
up-regulation motif.

\begin{figure*}[t]
\centerline{\subfigure{\includegraphics[trim = 5mm 15mm 10mm 0mm, clip,width=400pt]{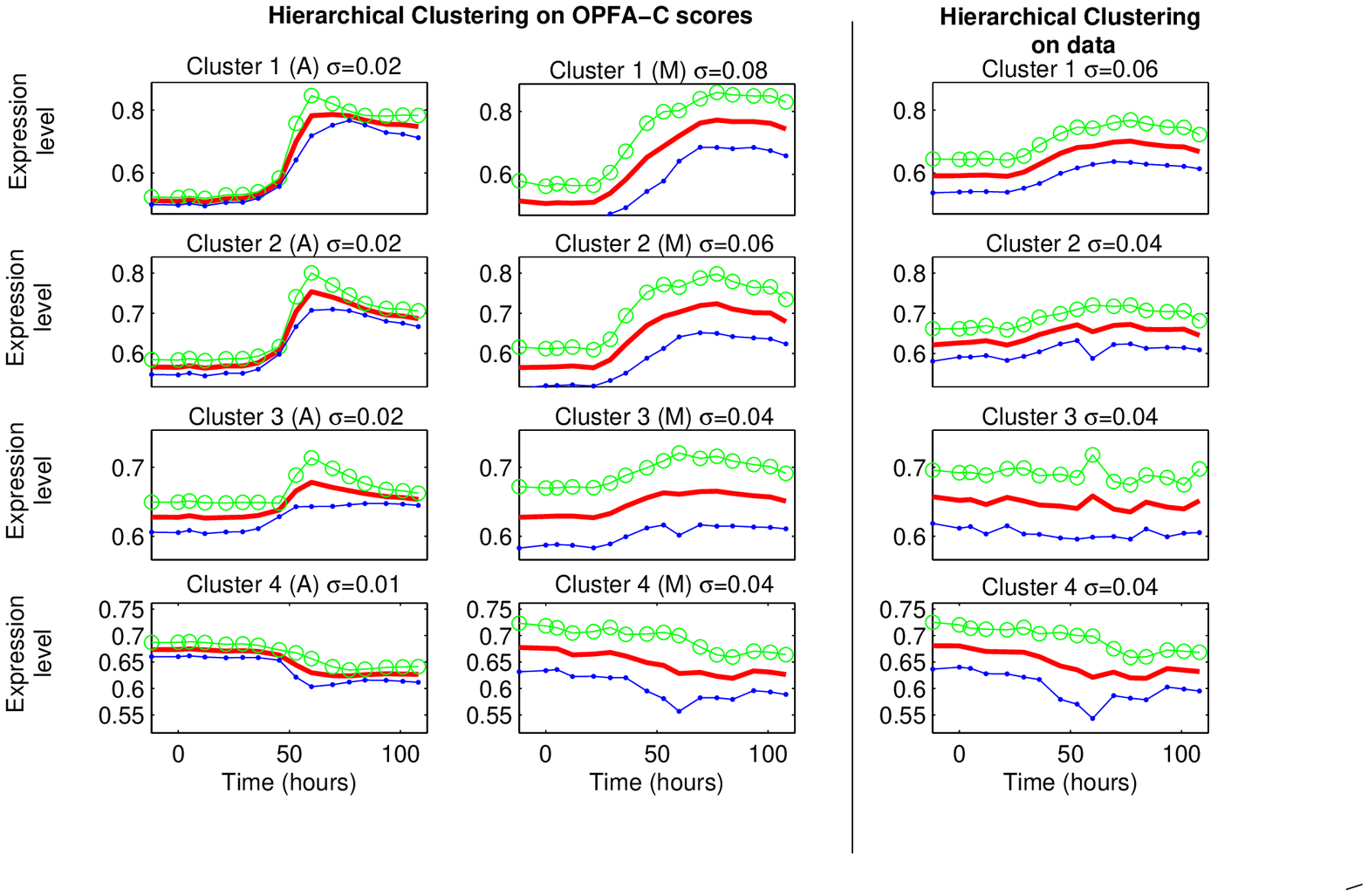}}}
\caption{The first two columns show the average expression signatures and their estimated upper/lower confidence intervals for each cluster of genes
  obtained by: averaging the \textit{estimated Aligned} expression
  patterns over the 
$S=9$ subjects (A) and directly averaging the
  misaligned observed data for each of the gene clusters obtained from the OPFA-C scores (M). 
The confidence intervals are computed according to $+ / -$ the estimated standard deviation 
at each time point. The cluster average standard deviation ($\sigma$) is computed
 as the average of the standard deviations at each time point.
  The last column shows the results of applying hierarchical clustering directly to the
original misaligned dataset $\left\{\b{X}_s\right\}_{s=1}^{S}$. In the first column, each  
 gene expression pattern is obtained by mixing the estimated aligned factors $\b{F}$ according to the estimated scores $\b{A}$. The alignment effect is
  clear, and interesting motifs become more evident.}
\label{Fig:expressionsignatures}
\end{figure*}

Finally, we consider the application of OPFA as a pre-processing step preceding a clustering analysis. 
Here the goal is to find groups of genes that share similar expression patterns and determine 
their characteristic expression patterns. In order to obtain gene clusters, we perform hierarchical clustering
 on the raw data ($\left\{\b{X}_s\right\}_{s=1}^{S}$) and on the lower
 dimensional space of the estimated factor scores ( $\{\b{A}_s\}_{s=1}^S$), 
obtaining two different sets of 4 well-differentiated clusters. We then compute the average expression
 signatures of the genes in each cluster by averaging over the observed data ($\left\{\b{X}_s\right\}_{s=1}^{S}$)
 and averaging over the data after OPFA correction for the temporal misalignments.  Figure \ref{Fig:expressionsignatures}
 illustrates the results. Clustering using the OPFA-C
factor scores produces a very significant improvement in cluster
concentration as compared to clustering using the raw data  $\left\{\b{X}_s\right\}_{s=1}^{S}$.
The first two columns in Figure compare the variation of the gene profiles over each
cluster for the temporally realigned data (labeled \textasciigrave A\textasciiacute) as compared to
to the profile variation of these same genes for the misaligned observed data
(labeled \textasciigrave M\textasciiacute).  For comparison, the last column shows
the results of applying hierarchical clustering directly to the
original misaligned dataset  $\left\{\b{X}_s\right\}_{s=1}^{S}$. It is clear that clustering on
the low-dimensional space of the OPFA-C scores unveils interesting motifs from the original noisy temporal expression trajectories.

\section{Conclusions}
\label{sec:conclusions}
We have proposed a general method of order-preserving factor analysis that
accounts for possible temporal misalignments in a population of
subjects undergoing a common treatment. We have described a simple
model based on circular-shift translations of prototype motifs and
have shown how to embed transient gene expression time courses into
this periodic model. The OPFA model can significantly improve
interpretability of complex misaligned data. The method is applicable
to other signal processing areas beyond gene expression time course
analysis.

A Matlab package implementing OPFA and OPFA-C will be available at the Hero Group Reproducible Research page (http://tbayes.eecs.umich.edu).

\appendix

\subsection{Circulant time shift model}

Using circular shifts in (\ref{factormodel}) introduces periodicity
into our model (\ref{originalstructure}). Some types of gene
expression may display periodicity, e.g. circadian transcripts, while
others, e.g. transient host response, may not. For transient gene
expression profiles such as the ones we are interested in here, we use
a truncated version of this periodic model, where we assume that each
subject's response arises from the observation of a longer periodic
vector within a time window (see Figure \ref{Fig:scheme}):
\begin{eqnarray}
\label{truncatedfactormodel}
\b{X}_s = \left[\b{M}\left(\b{F},\b{d}^s\right) \b{A}_s + \b{\epsilon}_s\right]_{\Omega}.
\end{eqnarray}
Here, the factors are of dimension $n_F \geq n$ and the window size is of dimension $n$ (in the special case that $n=n_F$, we have the original periodic model). In this model, the observed features are non-periodic as long as the delays $\b{d}^s$ are sufficiently small as compared to $n_F$. More concretely, if the maximum delay is $d_{\max}$, then in order to avoid wrap-around effects the dimension should be chosen as at least $n_F=n+ d_{\max}$. Finally, we  define the index set $\Omega$ corresponding to the observation window as:
\begin{eqnarray}
\label{Omegadef}
\Omega=\left\{\omega^1,\omega^1+1,\omega^1+2...,\omega^2\right\}^p
\end{eqnarray}
 where $\omega^1$ and $\omega^2$ are the lower and upper observation endpoints, verifying $n=\omega^2-\omega^1$.

\begin{figure}[htb]
\centerline{\subfigure{\includegraphics[trim = 5mm 20mm 10mm 20mm, clip,width=270pt]{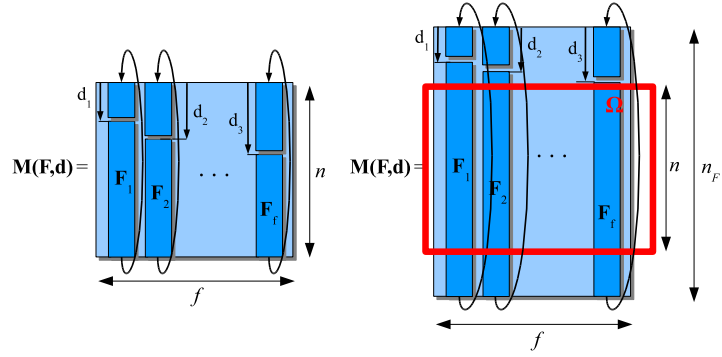}}}
\caption{\textit{Right}: Each subject's factor matrix $\b{M}_{i}$ is obtained by applying a circular shift to a common set of factors $\b{F}$ parameterized by a vector $\b{d}$. \textit{Left}: In order to avoid wrap-around effects when modeling transient responses, we consider instead a higher dimensional truncated model of dimension $n_F$ from which we only observe the elements within the window characterized by $\Omega$.}
\label{Fig:scheme}
\end{figure}

\subsection{Delay estimation and time-course alignment}

The solution to problem (\ref{JointProblem}) yields an estimate $\hat{\b{d}}^s$ for each subject's intrinsic factor delays. These delays are relative to the patterns found in the estimated factors and therefore require conversion to a common reference time. 

\begin{figure}[htb]
\centerline{\subfigure{\includegraphics[trim = 0mm 15mm 0mm 20mm, clip,width=250pt]{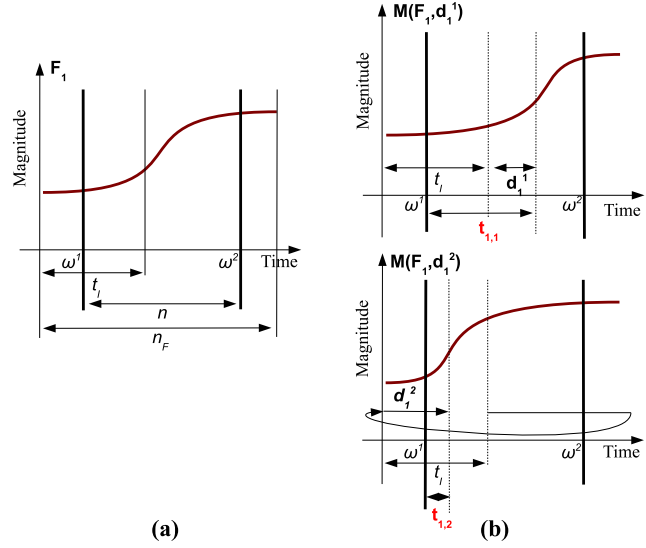}}}
\caption{(a) Time point of interest ($t_{I}$) for the up-regulation feature of factor 1. (b) The absolute time points $t_{1,1}$, $t_{1,2}$ are shown in red font for two different subjects and have been computed according to their corresponding relative delays and the formula in (\ref{timealignmentformula}).}
\label{fig:aligning}
\end{figure}

For a given up-regulation or down-regulation motif, $I$, which we call the feature of interest, found in factor $g$, we choose a time point of interest $t_{I}$. See Figure \ref{fig:aligning} (a) for an example of choice of $t_{I}$ for an up-regulation feature.

Then, given $t_{I}$ and for each subject $s$ and each factor $k$, we define the \textit{absolute feature occurrence time} as follows:

\begin{eqnarray}
\label{timealignmentformula}
t_{s,k} =\left(  \hat{\b{d}}^s_{k} + t_{I} \right) \mod ~ n_F - \omega^1.
\end{eqnarray}
where $\hat{\b{d}}^s_{g}$ is the estimated delay corresponding to factor $k$ and subject $s$ and $\omega^1$ is the lower endpoint of the observation window (see (\ref{Omegadef})).  
Figure \ref{fig:aligning} illustrates the computation of $t_{s,k}$ in a 2-factor example.

The quantities $t_{s,k}$ can be used for interpretation purposes 
or to \textit{realign} temporal profiles in order to find common, 
low-variance gene expression signatures, as shown in Section \ref{section:phddata}. 

\subsection{Implementation of $\mbox{EstimateFactors}$ and $\mbox{EstimateScores}$}
\label{Appendix1}

We consider here the implementation of $\mbox{EstimateFactors}$ and $\mbox{EstimateScores}$ 
under the presence of missing data. Let $\Omega_s=\left[\omega^s_1,\cdots,\omega^s_p \right] \in \left\{0,1 \right\}^{n\times p}$ 
be the set of observed entries in observation $\b{X}_s$. The objective in (\ref{JointProblem})
is then:
\begin{eqnarray}
\label{missingdataobjective}
\sum_{s=1}^S \l|\left[\b{X}_s - \b{M}\left({\b{F}},{\b{d}}^s \right){\b{A}}_s \right]_{\Omega_s}\r|_{F}^2
\end{eqnarray} 
We will show how to reformulate problems EstimateFactors (\ref{estimateFactors}) and  
EstimateScores (\ref{estimateScores}) in a standard quadratic objective with
linear and/or quadratic constraints.  

First, we rewrite the objective (\ref{missingdataobjective}) as:
\begin{eqnarray*}
\sum_{s=1}^S \sum_{j=1}^p \l|\diag\left(\omega^s_j\right) \left[\b{X}_s\right]_{\cdot,j} - \diag\left(\omega^s_j\right) \b{M}\left(\b{F},\b{d}^s\right)\left[\b{A}_s\right]_{\cdot,j} \r|_{F}^2.
\end{eqnarray*}
Expanding the square we obtain:
\begin{eqnarray}
\label{missingdataobjective2}
\sum_{s=1}^S \l|\left[\b{X}_s - \b{M}\left(\b{F},\b{d}^s\right)\b{A}_s\right]_{\Omega_s} \r|_F^2&=
\sum_{s=1}^S \sum_{j=1}^p \left[\b{A}_s\right]_{\cdot,j}^T \b{M}\left(\b{F},\b{d}^s\right)^T \diag\left(\omega^s_j\right) \b{M}\left(\b{F},\b{d}^s\right)\left[\b{A}_s\right]_{\cdot,j} \nonumber\\
&-2 \left[\b{X}_s\right]_{\cdot,j}^T\diag\left(\omega^s_j\right)\b{M}\left(\b{F},\b{d}^s\right)\left[\b{A}_s\right]_{\cdot,j} \nonumber\\
 & +\sum_{s=1}^S \l|\left[\b{X}_s\right]_{\Omega_s} \r|_F^2.
\end{eqnarray}

To obtain the EstimateFactors objective, first we will rewrite the OPFA model (\ref{truncatedfactormodel})
 using matrix notation. Let $\b{U}_i$ be a circular shift matrix $\b{U}_i$ parameterized by the $i$-th delay $\b{d}$ component. Then
\begin{eqnarray*}
\b{M}\left(\b{F},\b{d}\right)=\left[\b{U}_1\b{F}_{1},\cdots, \b{U}_f\b{F}_{f}\right]=\b{H}\tilde{\b{F}}
\end{eqnarray*}
where $\b{F}_{j}$ denotes the $j$-th column of $\b{F}$, $\b{H}$ is the concatenation of the $\b{U}_i$ 
matrices and $\tilde{\b{F}}$ is a matrix containing the columns of $\b{F}$ with the appropriate padding of zeros.
With this notation and (\ref{missingdataobjective2}) we obtain:
\begin{eqnarray}
\sum_{s=1}^S \l|\left[\b{X}_s - \b{M}\left(\b{F},\b{d}^s\right)\b{A}_s\right]_{\Omega_s} \r|_F^2& \propto_{\b{F}}
\sum_{s=1}^S \sum_{j=1}^p  \trace\left( \left[\b{A}_s\right]_{\cdot,j}\left[\b{A}_s\right]_{\cdot,j}^T \tilde{\b{F}}^T \b{H}_s^T\diag\left(\omega^s_j\right) \b{H}_s\tilde{\b{F}} \right) \nonumber\\
&-2 \trace\left(\left[\b{A}_s\right]_{\cdot,j} \left[\b{X}_s\right]_{\cdot,j}^T\diag\left(\omega^s_j\right)\b{H}_s\tilde{\b{F}}\right)\nonumber.
\end{eqnarray}
We now use the identity (\cite{neudecker1999matrix}, Thm. 3 Sec. 4):
\begin{eqnarray}
\label{traceidentity}
\trace\left(\b{Z}\b{X}^T\b{Y}\b{W}\right)=\vec\left(\b{W}\right)^T\b{Z}\otimes \b{Y}^T\vec\left(\b{X}\right)\nonumber,
\end{eqnarray}
to write:
\begin{eqnarray}
\sum_{s=1}^S \l|\left[\b{X}_s - \b{M}\left(\b{F},\b{d}^s\right)\b{A}_s\right]_{\Omega_s} \r|_F^2& \propto_{\b{F}}
 \vec\left( \tilde{\b{F}}\right)^T \left(\sum_{s=1}^S \sum_{j=1}^p \left[\b{A}_s\right]_{\cdot,j}\left[\b{A}_s\right]_{\cdot,j}^T \otimes  \b{H}_s^T \diag\left(\omega^s_j\right)\b{H}_s\right) \vec \left(\tilde{\b{F}} \right) \nonumber\\
&-2 \vec\left( \sum_{s=1}^S \sum_{j=1}^p  \b{H}_s^T \diag\left(\omega^s_j\right)\left[\b{X}_s\right]_{\cdot,j}\left[\b{A}_s\right]_{\cdot,j}^T\right) \vec \left(\tilde{\b{F}} \right)  \nonumber.
\end{eqnarray}
Finally, making use of the fact that $\tilde{\b{F}}$ is a block-column matrix with the columns of $\b{F}$ padded by zeros,
we conclude:
\begin{eqnarray}
\sum_{s=1}^S \l|\left[\b{X}_s - \b{M}\left(\b{F},\b{d}^s\right)\b{A}_s\right]_{\Omega_s} \r|_F^2& \propto_{\b{F}}
 \vec\left(\b{F}\right)^T \b{Q}_{F} \vec \left({\b{F}} \right) -2 \b{q}_{F}^T\vec \left({\b{F}} \right)  \nonumber
\end{eqnarray}
where we have defined
\begin{eqnarray}
\b{Q}_{F}=\left[\sum_{s=1}^S \sum_{j=1}^p \left[\b{A}_s\right]_{\cdot,j}\left[\b{A}_s\right]_{\cdot,j}^T \otimes  \b{H}_s^T \diag\left(\omega^s_j\right)\b{H}_s\right]_{\mathcal{J},\mathcal{J}}\\
\b{q}_{F}=\left[\vec\left( \sum_{s=1}^S \sum_{j=1}^p  \b{H}_s^T \diag\left(\omega^s_j\right)\left[\b{X}_s\right]_{\cdot,j}\left[\b{A}_s\right]_{\cdot,j}^T\right) \right]_{\mathcal{J}}
\end{eqnarray}
and $\mathcal{J}$ are the indices corresponding to the non-zero elements in $\vec\left( \tilde{\b{F}}\right)$.
Hence, EstimateFactors can be written as:
\begin{eqnarray*}
\min_{\b{F}} &  \vec\left(\b{F}\right)^T \left( \b{Q}_{F} +  \beta \diag\left(\b{W}^T\b{W},\cdots, \b{W}^T\b{W} \right)\right)\vec \left({\b{F}} \right) -2 \b{q}_{F}^T\vec \left({\b{F}} \right) \\ 
\mbox{ s.t. } & \left\{
\begin{array}{c c}
 \l| \b{F} \r|^2_{\mathcal{F}} \leq \delta \nonumber & \\
 \b{F}_{i,j} \geq 0 & i=1,\cdots,n,\\
&  j=1,\cdots,f
\end{array}
\right.
\end{eqnarray*}
The dimension of the variables in this problem is $n f$. In the applications considered
here, both $n$ and $f$ are relatively small and hence this program can be solved with a standard 
convex solver such as SeDuMi \cite{sturm1999using} (upon conversion to a standard conic problem).

On the other hand, we can follow the same procedure to reformulate  the objective 
in EstimateScores (\ref{estimateScores}) into a penalized quadratic form. First we use
(\ref{missingdataobjective2}) and (\ref{traceidentity}) to write:
\begin{eqnarray}
\sum_{s=1}^S \l|\left[\b{X}_s - \b{M}\left(\b{F},\b{d}^s\right)\b{A}_s\right]_{\Omega_s} \r|_F^2&\propto_{\left\{\b{A}_s\right\}_{i=1}^S}
\sum_{s=1}^S \vec\left(\b{A}_s\right)^T \b{Q}^s_{A}  \vec\left(\b{A}_s\right)-2{\b{q}^s_{A}}^T\vec\left(\b{A}_s\right)  \nonumber
\end{eqnarray}
where
\begin{eqnarray}
\b{Q}^s_{A} = \left[\begin{array}{cccc}
 \b{M}\left(\b{F},\b{d}^s\right)^T \diag\left(\omega^s_1\right) \b{M}\left(\b{F},\b{d}^s\right)  & 
\b{0} &\dots&  \b{0}\\
 \b{0} &\b{0}&  \cdots & \b{0}         \\        
  \b{0} &\cdots&  \b{0} & \b{M}\left(\b{F},\b{d}^s\right)^T \diag\left(\omega^s_p\right) \b{M}\left(\b{F},\b{d}^s\right)    
\end{array}\right]\\
\b{q}^s_{A} = \left[\begin{array}{ccc}
\left[\b{X}_s\right]_{\cdot,1}^T\diag\left(\omega^s_1\right)\b{M}\left(\b{F},\b{d}^s\right) & \cdots &
 \left[\b{X}_s\right]_{\cdot,p}^T\diag\left(\omega^s_p\right)\b{M}\left(\b{F},\b{d}^s\right)
\end{array}\right].
\end{eqnarray}
Thus EstimateFactors can be written as:
\begin{eqnarray}
\label{estimateFactorsMD}
\min_{\b{F}} &  \sum_{s=1}^S \vec\left(\b{A}_s\right)^T \b{Q}^s_{A}  \vec\left(\b{A}_s\right)-2{\b{q}^s_{A}}^T\vec\left(\b{A}_s\right) + \lambda \sum_{i=1}^p\sum_{j=1}^f \|\left[\b{A}_1\right]_{j,i} \cdots \left[\b{A}_S \right]_{j,i}\|_2\\ 
\mbox{ s.t. } & \left\{
\begin{array}{c c}
 \l| \b{F} \r|^2_{\mathcal{F}} \leq \delta \nonumber & \\
 \b{F}_{i,j} \geq 0 & i=1,\cdots,n,\\
&  j=1,\cdots,f
\end{array}
\right.
\end{eqnarray}

This is a convex, non-differentiable and potentially high-dimensional problem.
 For this type of optimization problems, there exists a class of simple and 
scalable algorithms which has recently received much attention \cite{Zibulevsky2010},
 \cite{daubechies2003iterative}, \cite{combettes2006signal}, \cite{pustelnikconstrained}.
These algorithms rely only on first-order updates of the type:
\begin{eqnarray}
\label{FBalgo}
\b{\b{x}}^t\leftarrow \mathcal{T}_{\Gamma, \lambda}\left(\b{v} - 2\b{\b{x}}^{t-1}\left( \alpha\b{I} -\b{Q}\right)\right), 
\end{eqnarray}
which only involves matrix-vector multiplications and evaluation of the operator $\mathcal{T}$,
which is called the proximal operator \cite{combettes2006signal} associated to $\Gamma$ and $\mathcal{C}$ and is defined as:
\begin{eqnarray}
\label{Operator}
\mathcal{T}_{\Gamma, \lambda}\left(\b{v}\right)&:=\min & \frac{1}{2}\b{x}'\b{x}+ \b{v}'\b{x} + \lambda \Gamma\left(\b{x}\right) \nonumber \\ 
~ & \mbox{ s.t. } & \b{x}  \in \mathcal{C}  \nonumber.
\end{eqnarray}
This operator takes the vector $\b{v}$ as an input and outputs a shrunk/thresholded version 
of it depending on the nature of the penalty $\Gamma$ and the constraint set $\mathcal{C}$. 
For some classes of penalties $\Gamma$ (e.g. $l_1$, $l_2$, mixed $l_{1}-l_{2}$)
 and the positivity constraints considered here, this operator has a closed 
form solution \cite{tibaupuig2009}, \cite{combettes2006signal}. 
Weak convergence of the sequence (\ref{FBalgo}) to the optimum of (\ref{estimateFactorsMD}) is assured 
for a suitable choice of the constant $\alpha$ \cite{pustelnikconstrained}, \cite{Beck2008}.

\subsection{Delay Estimation lower bound in the presence of Missing Data}
\label{Appendix2}
As we mentioned earlier, the lower bound (\ref{LB}) does not hold anymore under the presence of missing data. 
We derive here another bound that can be used in such case. 
From expression (\ref{missingdataobjective2}), we first obtain the objective in
 EstimateDelays (\ref{EstimateDelayproblem}) in a quadratic form:
\begin{eqnarray*}
\sum_{s=1}^S \l|\left[\b{X}_s - \b{M}\left(\b{F},\b{d}^s\right)\b{A}_s\right]_{\Omega_s} \r|_F^2&\propto_{\b{d}^s} 
\sum_{s=1}^S \sum_{j=1}^p \trace\left(\b{M}\left(\b{F},\b{d}^s\right)^T \diag\left(\omega^s_j\right) \b{M}\left(\b{F},\b{d}^s\right)\b{P}_{s,j}\right)\\
&-2 \trace\left(\b{Q}_{s,j}\b{M}\left(\b{F},\b{d}^s\right)\right).
\end{eqnarray*}
Where we have let $\b{P}_{s,j}:= \left[\b{A}_s\right]_{\cdot,j}\left[\b{A}_s\right]_{\cdot,j}^T$
 and $\b{Q}_{s,j}=\left[\b{A}_s\right]_{\cdot,j}\left[\b{X}_s\right]_{\cdot,j}^T\diag\left(\omega^s_j\right)$. Notice that each 
of the terms indexed by $s$ is independent of the others. Using (\ref{traceidentity}), we obtain
\begin{eqnarray*}
 \l|\left[\b{X}_s - \b{M}\left(\b{F},\b{d}^s\right)\b{A}_s\right]_{\Omega_s} \r|_F^2 & \propto 
\vec\left(\b{M}\left(\b{F},\b{d}^s\right)\right)^T\sum_{j=1}^p \left( \b{P}_{s,j}\otimes \diag\left(\omega^s_j\right)\right)\vec\left(\b{M}\left(\b{F},\b{d}^s\right)\right) \\
& -2 \vec\left(\sum_{j=1}^p \b{Q}_{s,j}^T\right)^T \vec\left(\b{M}\left(\b{F},\b{d}^s\right)\right).
\end{eqnarray*}
We now can minorize the function above by:
\begin{eqnarray*}
 \l|\left[\b{X}_s - \b{M}\left(\b{F},\b{d}^s\right)\b{A}_s\right]_{\Omega_s} \r|_F^2 &\geq  
\underline{\lambda}\left(\sum_{j=1}^p  \b{P}_{s,j}\otimes \diag\left(\omega^s_j\right)\right)\vec\left(\b{M}\left(\b{F},\b{d}^s\right)\right)^T\vec\left(\b{M}\left(\b{F},\b{d}^s\right)\right) \\
& -2 \vec\left(\sum_{j=1}^p \b{Q}_{s,j}^T\right)^T \vec\left(\b{M}\left(\b{F},\b{d}^s\right)\right)+\sum_{s=1}^S \l|\left[\b{X}_s\right]_{\Omega_s} \r|_F^2
\end{eqnarray*}
which leads to:
\begin{eqnarray*}
 \l|\left[\b{X}_s - \b{M}\left(\b{F},\b{d}^s\right)\b{A}_s\right]_{\Omega_s} \r|_F^2 &\geq  
\underline{\lambda}\left(\sum_{j=1}^p  \b{P}_{s,j}\otimes \diag\left(\omega^s_j\right)\right) \l|\b{F} \r|_F^2\\
& -2 \vec\left(\sum_{j=1}^p \b{Q}_{s,j}^T\right)^T \vec\left(\b{M}\left(\b{F},\b{d}^s\right)\right)+\l|\left[\b{X}_s\right]_{\Omega_s} \r|_F^2.
\end{eqnarray*}
Using the relaxation in (\ref{LBRelaxation}), we can now compute a lower bound $\Phi_{lb}\left({\mathcal{I}}_t\right)$
as
\begin{eqnarray*}
\Phi_{lb}\left({\mathcal{I}}_t\right) & =  
\underline{\lambda}\left(\sum_{j=1}^p  \b{P}_{s,j}\otimes \diag\left(\omega^s_j\right)\right) \l|\b{F} \r|_F^2 +\l|\left[\b{X}_s\right]_{\Omega_s} \r|_F^2 \\
& -2 \max_{\b{d}\in {\mathcal{S}}_t}\vec\left(\sum_{j=1}^p \b{Q}_{s,j}^T\right)^T \vec\left(\b{M}\left(\b{F},\b{d}^s\right)\right).
\end{eqnarray*}

\bibliographystyle{IEEEtran}
\bibliography{OrdinalDictionaryLearning}
\end{document}